\documentclass{article}

\usepackage[final]{neurips_2021}

\usepackage{microtype}
\usepackage{graphicx}
\usepackage{subcaption}
\usepackage{booktabs} % for professional tables
\usepackage{algorithm} 
\usepackage[noend]{algcompatible}
\usepackage{wrapfig} 

\include{Definitions}
\graphicspath{ {./figures/} }
\usepackage[dvipsnames]{xcolor}

\usepackage{thmtools,thm-restate}

\newif\ifcomments

\commentstrue      % Show comments
\title{Continual Auxiliary Task Learning}

\author{
  Matthew McLeod, Chunlok Lo, Matthew Schlegel, Andrew Jacobsen, Raksha Kumaraswamy\\
  Department of Computing Science, University of Alberta\\
  \{\texttt{mmcleod2,chunlok,mkschleg,ajjacobs,kumarasw}\}\texttt{@ualberta.ca}
  \And
  Martha White, Adam White\\
   Department of Computing Science, University of Alberta\\
   CIFAR Canada AI Chair, Alberta Machine Intelligence Institute (Amii)\\
  \{\texttt{whitem,amw8}\}\texttt{@ualberta.ca} 
  \vspace{-0.2cm}
}

\begin{document}

\maketitle

\begin{abstract}
Learning auxiliary tasks, such as multiple predictions about the world, can provide many benefits to reinforcement learning systems. A variety of off-policy learning algorithms have been developed to learn such predictions, but as yet there is little work on how to adapt the behavior to gather useful data for those off-policy predictions. In this work, we investigate a reinforcement learning system designed to learn a collection of auxiliary tasks, with a behavior policy learning to take actions to improve those auxiliary predictions. We highlight the inherent non-stationarity in this continual auxiliary task learning problem, for both prediction learners and the behavior learner. We develop an algorithm based on successor features that facilitates tracking under non-stationary rewards, and prove the separation into learning successor features and rewards provides convergence rate improvements. We conduct an in-depth study into the resulting multi-prediction learning system. 
\end{abstract}
% and demonstrate learning failures with standard algorithms that assume stationarity, like those that use replay. 
%To the best of our knowledge, no work has yet investigated a full multi-prediction reinforcement learning system, where both the behavior agent and prediction learners are designed so that the whole collection of predictions can be learned efficiently. Such a multi-prediction system has been recently investigated in the bandit setting  \cite{DBLP:journals/corr/LinkeCoRR2019}, with a thorough empirical study of different intrinsic rewards to drive behavior and different prediction learners. The work proposed the use of intrinsic rewards from learners as the balancing mechanism between prediction learners. Furthermore, it emphasized the importance of using learners that can modulate their own learning rate based on whether progress can be made. The insights there inform this work, but they do not handle the unique challenges of a state dependent setting. It did not handle the dependency of prediction on state, the long-term consequences of actions, and the sharing of samples between prediction learners, as each prediction (arm) in the bandit setting is independent with no generalization between prediction learners possible. The goal of this work is to investigate how to develop an efficient reinforcement learning system in a state dependent environment, where the primary goal is to learn and maintain multiple predictions that are possibly non-stationary.

\section{Introduction}
\label{sec:introduction}

In never-ending learning systems, the agent often faces long periods of time when the external reward is uninformative. A smart agent should use this time to practice reaching subgoals, learning new skills, and refining model predictions. Later, the agent should use this prior learning to efficiently maximize external reward. The agent engages in this self-directed learning during times when the primary drives of the agent (e.g., hunger) are satisfied. Other times, the agent might have to trade-off directly acting towards internal auxiliary learning objectives and taking actions that maximize reward. 

% MARTHAC: I am not yet sure how to incorporate this. The main message here is: even though there is a trade-off, a key component is taking actions to learning auxiliary tasks. But maybe this need not be said yet, as the rest of the intro motivates it
%Recent research has shown that agents that incorporate auxiliary learning objectives, but select actions only to optimize the main task, surprisingly outperform agents without auxiliary learning objectives. For example, Deep Q-learning agents appear to learn better representations (and perform better in Atari) when the network is updated in service of both the value function and additional auxiliary value functions \cite{jaderberg2016reinforcement}. This {\em auxiliary task effect} correlates with the agent's progress on all of its learning objectives---inaccurate value functions or mostly random policies will likely slow main task learning. Clearly the agent's action selection should focus at least occasionally on improving the axillary tasks.

In this paper we investigate how an agent should select actions to balance the needs of several auxiliary learning objectives in a {\em no-reward setting} where no external reward is present. In particular, we assume the agent's auxiliary objectives are to learn a diverse set of value functions corresponding to a set of fixed policies. Our solution at a high-level is straightforward. Each auxiliary value function is learned in parallel and off-policy, and the behavior selects actions to maximize learning progress. Prior work investigated similar questions in a state-less bandit like setting, where both off-policy learning and function approximation are not required ~\citep{linke2020adapting}. 

%MARTHAC: Maybe it is too soon to say this
%Within this framework, however, many challenges arise. Some of the auxiliary learning problems may be non-stationary, and thus the agent must continually select actions to keep the predictions up to date. Other tasks are easy to learn, and can be largely ignored after some amount of time. Ideally, the behavior will take actions that lead to new data that is useful for many learners, without starving a prediction learner that requires samples in other parts of the environment. Obtaining an effective learning system requires sample efficient off-policy prediction learners that can learn under a changing behavior and learning a behavior policy that adapts under constantly changing learning progress. 

% TODO: Does work on skills and transfer fit anywhere here, like konidaris2007building
Otherwise, the majority of prior work has focused on how the agent could make use of auxiliary learning objectives, not how behavior could be used to improve auxiliary task learning. 
 Some work has looked at defining (predictive) features, such as successor features and a basis of policies \citep{barreto2018transfer,borsa2018universal,barreto2020fast,NEURIPS2019_251c5ffd}; universal value function approximators \citep{SchaulICML2015}; and features based on value predictions \citep{schaul2013better,schlegel2021general}. 
 %These approaches all assumed simple fixed behaviors or randomly collected datasets. 
 The other focus has been exploration, using auxiliary learning objectives to generate bonuses to aid exploration on the main task \citep{pathak2017curiosity,stadie2015incentivizing,puigdomenech2020never,burda2018exploration}; using a given set of policies in a call-return fashion for scheduled auxiliary control \citep{riedmiller2018learning}; and discovering subgoals in environments where it is difficult for the agent to reach particular parts of the state-action space \citep{machado2017laplacian,colas2019curious,zhang2020generating,andrychowicz2017hindsight,pong2019skew}. In all of these works, the behavior was either fixed or optimized for the main task. 

The problem of adapting the behavior to optimize many auxiliary predictions in the absence of external reward is sufficiently complex to merit study in isolation. It involves several inter-dependent learning mechanisms, multiple sources of non-stationarity, and high-variance due to off-policy updating. If we cannot design learning systems that efficiently learn their auxiliary objectives in isolation, then the agent is unlikely to learn its auxiliary tasks while additionally balancing external reward maximization. 

Further, understanding how to efficiently learn a collection of auxiliary objectives is complementary to the goals of using those auxiliary objectives. It could amplify the auxiliary task effect in UNREAL \citep{jaderberg2016reinforcement}, improve the efficiency and accuracy of learning successor features and universal value function approximators, and improve the quality of the sub-policies used in scheduled auxiliary control. It can also benefit the numerous systems that discover options, skills, and subgoals ~\citep{gregor2016variational,eysenbach2018diversity,veeriah2019discovery,pitis2020maximum,nair2020contextual,pertsch2020long,colas2019curious,eysenbach2019search}, by providing improved algorithms to learn the resulting auxiliary tasks. For example, for multiple discovered subgoals, the agent can adapt its behavior to efficiently learn policies to reach each subgoal. 
% MARTHAC: Maybe too nuanced here? Maybe this is better put in a conclusion, actually, saying what could come next
%In addition, the approach developed in this paper can be used to generalize single-task exploration methods (e.g., \cite{pathak2017curiosity}) to balance multiple intrinsic reward bonuses, instead of one.    

In this paper we introduce an architecture for parallel auxiliary task learning. As the first such work to tackle this question in reinforcement learning with function approximation, numerous algorithmic challenges arise. 
%Of particular importance is the inherent non-stationarity in the problem, which precludes the straightforward use of most reinforcement learning algorithms. 
We first formalize the problem of learning multiple predictions as a reinforcement learning problem, and highlight that the rewards for the behavior policy are inherently non-stationary due to changes in learning progress over time. We develop a strategy to use successor features to exploit the stationarity of the dynamics, whilst allowing for fast tracking of changes in the rewards, and prove that this separation provides a faster convergence rate than standard value function algorithms like temporal difference learning. We empirically show that this separation facilitates tracking both for prediction learners with non-stationary targets as well as the behavior. 

\section{Problem Formulation}

We consider the \emph{multi-prediction problem}, in which an agent continually interacts with an environment to obtain accurate predictions. This interaction is formalized as a Markov decision process (MDP), defined by a set of states $\States$, a set of actions $\Actions$, and a transition probability function $ \Prob(s, a, s')$. The agent's goal, when taking actions, is to gather data that is useful for learning $\numpred$ predictions, where each prediction corresponds to a general value function (GVF) \citep{sutton2011horde}.

A GVF question is formalized as a three tuple $(\pi, \gamma, \cfunc)$, where the target is the expected return of the cumulant, defined by $\cfunc: \States \times \Actions \times \States \rightarrow \RR$, when following policy $\pi: \States \times \Actions \rightarrow [0,1]$, discounted by $\gamma: \States \times \Actions \times \States \rightarrow [0,1]$. More precisely, the target is the action-value
\begin{align*}
Q(s, a) &\defeq \mathbb{E}_{\pi}\left[G_t | S_t = s, A_t = a\right] \quad
\text{ for } G_t \defeq C_{t+1} + \gamma_{t+1} G_{t+1} 
\end{align*}
where $C_{t+1} \defeq \cfunc(S_t, A_t, S_{t+1})$ and $\gamma_{t+1} \defeq \gamma(S_t, A_t, S_{t+1})$. The extension of $\gamma$ to transitions allows for a broader class of problems, including easily specifying termination, without complicating the theory \citep{white2017unifying}. The expectation is under policy $\pi$, with transitions according to $\Prob$. The prediction targets could also be state-value functions; we assume the targets are action-values in this work to provide a unified discussion of successor features for both the GVF and behavior learners.

At each time step, the agent produces $\numpred$ predictions, a $\Qhat_t^{(\predind)}(S_t, A_t)$ for prediction $\predind$ with true targets $Q^{(\predind)}_t(S_t, A_t)$. We assume the GVF question can change over time, and so $Q$ can change with time. The goal is to have low error in the prediction, in terms of the root mean-squared value error (RMSVE), under state-action weighting $d: \States \times \Actions \rightarrow \RR$:
\begin{equation}
\RMSVE(\Qhat, Q) \defeq \sqrt{\sum_{s \in \States} \sum_{a \in \Actions} d(s, a) (\hat{Q}(s, a) - Q(s, a))^2}
\end{equation}
The total error up to time step $t$, across all predictions, is $\TE \ \defeq  \sum_{\timeind=1}^t \sum_{\predind=1}^\numpred \RMSVE(\Qhat_\timeind^{(\predind)}, Q^{(\predind)}_\timeind)$.
%\begin{equation}
%\TE \ \defeq  \sum_{\timeind=1}^t \sum_{\predind=1}^\numpred \RMSVE(\Qhat_\timeind^{(\predind)}, Q^{(\predind)}_\timeind)
%.
%\end{equation}
%%

\begin{wrapfigure}[10]{r}{0.24\textwidth}
\centering
%\vspace{-0.1cm}
 	\includegraphics[width=0.25\textwidth]{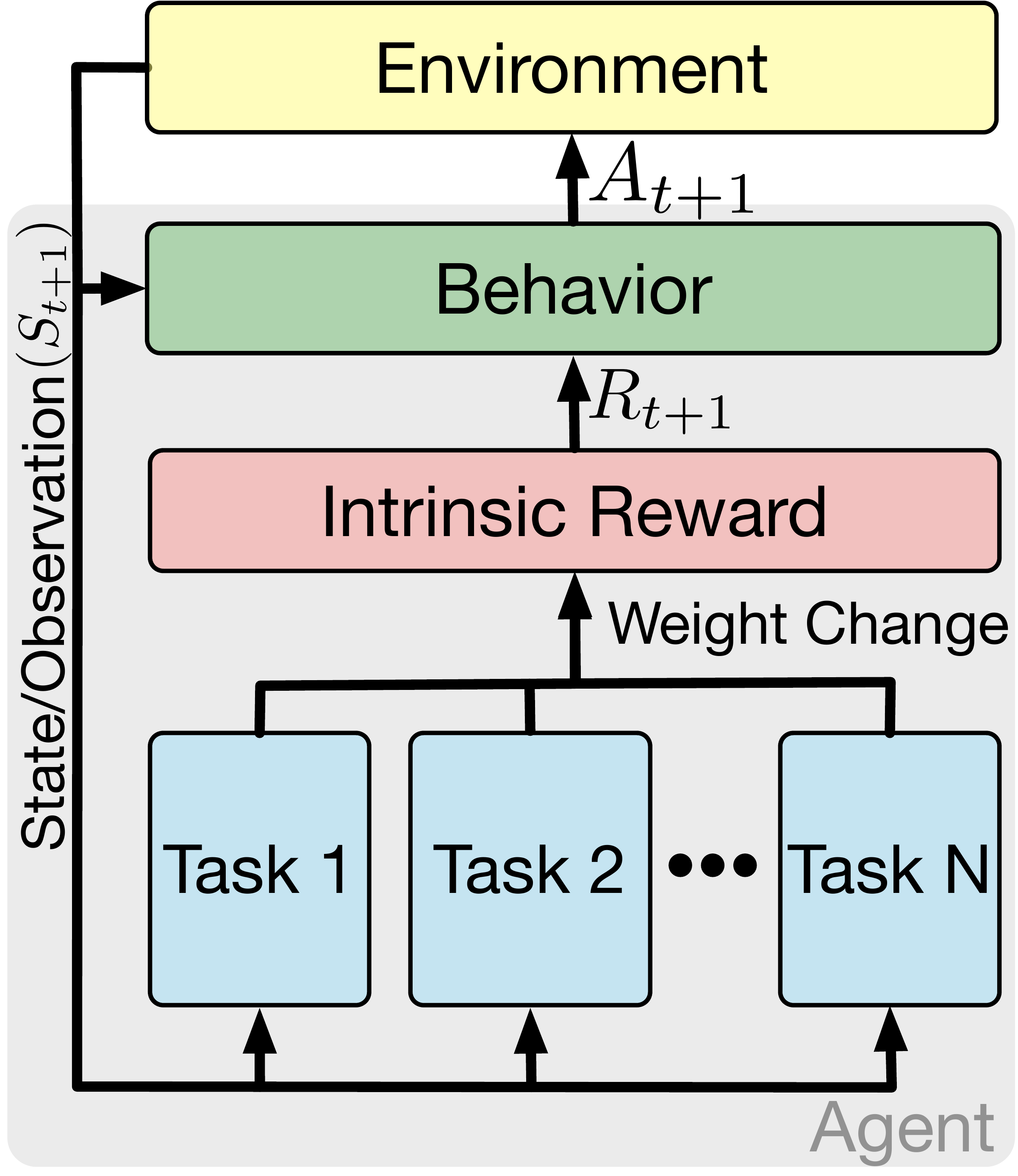}
 	%\caption{Using interest functions. Shading is standard error over 30 runs.}
 	\label{fig:system}
\end{wrapfigure}
The agent's goal is to gather data and update its predictions to make \TE ~small. This goal can itself be formalized as a reinforcement learning problem, by defining rewards for the behavior policy that depend on the agent's predictions. Such rewards are often called \emph{intrinsic rewards}. For example, if we could directly measure the RMSVE, one potential intrinsic reward would be the decrease in the RMSVE after taking action $A_t$ from state $S_t$ and transitioning to $S_{t+1}$. This reflects the agent's learning progress---how much it was able to learn---due to that new experience. The reward is high if the action generated data that resulted in substantial learning. While the RMSVE is the most direct measure of learning progress, it cannot be calculated without the true values. 
%one cannot calculate it without access to the true GVF.

\begin{wrapfigure}[19]{r}{0.45\textwidth}
\vspace{-0.7cm}
\begin{minipage}{0.48\textwidth}
 \begin{algorithm}[H]
   \caption{Multi-Prediction Learning System}
   \label{alg:generic}
    {\bfseries Input:} $\numpred$ GVF questions
 \begin{algorithmic}
   \STATE {Initialize behavior policy parameters } $\bweights_0$\\ \text{ and GVF learners } $w^{(1)}_{0}, \ldots, w^{(\numpred)}_0$
   \STATE {Obtain initial observation } $S_0$
   \FOR{$t = 0, 1, \ldots$}
   \STATE Choose action $A_t$ according to $\pi_{\bweights_t}(\cdot | S_t)$
  \STATE Observe next state vector $S_{t+1}$ and $\gamma_{t+1}$
  \STATE // Update predictions with new data
   \FOR{$\predind=1$ {\bfseries to} $\numpred$}
       \STATE $c \gets \cfunc^{(\predind)}(S_t, A_t, S_{t+1})$
       \STATE $\gamma \gets \gamma^{(\predind)}(S_t, A_t, S_{t+1})$
      \STATE Update $w_t^{(j)}$ with $(S_t, A_t, c, S_{t+1}, \gamma)$
   \ENDFOR
  \STATE \!\!\!// Compute intrinsic reward, update behavior
   \STATE $R_{t+1} \gets \sum_{j=1}^N \| w^{(\predind)}_{t+1} - w^{(\predind)}_{t} \|_1$
%   \FOR{$\predind=1$ {\bfseries to} $\numpred$}
 %      \STATE $r \gets \sum_{j=1}^N r + \| w^{(\predind)}_{t+1} - w^{(\predind)}_{t} \|_1$
  % \ENDFOR
   \STATE Update $\bweights_t$ with $(S_t, A_t, R_{t+1}, S_{t+1}, \gamma_{t+1})$ 
   \ENDFOR
 \end{algorithmic}
 \end{algorithm}
\end{minipage}
\end{wrapfigure}
Many intrinsic rewards have been considered to estimate the learning progress of predictions. A recent work provided a thorough survey of different options, as well as an empirical study \citep{linke2020adapting}. Their conclusion was that, for reasonable prediction learners, simple learning progress measures---like the change in weights---were effective for producing effective data gathering. We rely on this conclusion here, and formalize the problem using the $\ell_1$ norm on the change in weights. Other intrinsic rewards could be swapped into the framework; but, because our focus is on the non-stationarity in the system and because empirically we found this weight-change intrinsic reward to be effective, we opt for this simple choice upfront.
%to avoid exploring all dimensions of the multi-prediction problem in reinforcement learning.

We provide the generic pseudocode for a multi-prediction reinforcement learning system, in Algorithm \ref{alg:generic}. Note that the behavior agent also has a separate transition-based $\gamma$, which enables us to encode both continuing and episodic problems. For example, the pseudo-termination for a GVF could be a particular state in the environment, such as a doorway. The discount for the GVF would be zero in that state, even though it is not a true terminal state; the behavior discount $\gamma_{t+1}$ would not be zero.

\section{Non-stationarity Induced by Learning}
\label{sec:nonstationarity-in-learning}

On the surface, the multi-prediction problem outlined in the previous section is a relatively straightforward reinforcement learning problem. The behavior policy learns to maximize cumulative reward, and simultaneously learns predictions about its environment. Many RL systems incorporate prediction learning, either as auxiliary tasks or to learn a model. However, unlike standard RL problems, the rewards for the behavior are non-stationary when using intrinsic rewards, even under stationary dynamics. Further, the prediction problems themselves are non-stationary due to a changing behavior.
% policy, as the prediction loss depends on the behaviour state distribution.

To understand this more deeply, consider first the behavior rewards. On each time step, the predictions are updated. Progressively, they get more and more accurate. 
Imagine a scenario where they can become perfectly accurate, such as in the tabular setting with stationary cumulants. 
The behavior rewards are high in early learning, when predictions are inaccurate. As predictions become more and more accurate, the change in weights gets smaller until eventually the behavior rewards are near zero. This means that when the behavior revisits a state, the reward distribution has actually changed.
More generally, in the function approximation setting, the behavior rewards will continue to change with time, not necessarily decay to zero. 
% but not necessarily by decaying to zero. We can often expect the best realizable estimated value function to change over time, due to tracking. The function approximator for the predictions is limited, and so the weights may adjust to specialize to more recently observed parts of the environment. Local learning progress is therefore constantly occurring, as the agent improves its estimates for the current region of the environment. 

The prediction problems are also non-stationary for two reasons.
First, the cumulants themselves might be non-stationary, even if the transition dynamics are stationary. For example, the cumulant could correspond to the amount of food in a location in the environment, that slowly gets depleted. Or, the cumulant could depend on a hidden variable, that makes the outcome appear non-stationary.
% MARTHAC: This sentence seems out of place  
%The overall conclusion is that we need an algorithm for the behavior that accounts for this non-stationarity.
%
Even with a stationary cumulant, the prediction learning problem can be non-stationary due to a changing behavior policy. As the behavior policy changes, the state distribution changes. Implicitly, when learning off-policy, the predictions are minimizing an objective weighted by the state visitation under the behavior policy. As the behavior changes, the underlying objective is actually changing, resulting in a non-stationary prediction problem. 
%Unlike the inherent non-stationarity in the behavior rewards, however, we can actually control this source of non-stationarity by explicitly re-weighting states. We discuss this algorithmic modification in the next section.

Though there has been some work on learning under non-stationarity in RL and bandits, none to our knowledge has addressed the multi-prediction setting in MDPs. There has been some work developing reinforcement learning algorithms for non-stationary MDPs, but largely for the tabular setting \citep{suttonbartobook,da2006improving,abdallah2016addressing,cheung2020reinforcement} or assuming periodic shifts \citep{chandak2020towards,chandak2020optimizing,padakandla2020reinforcement}. There has also been some work in the non-stationary multi-armed bandit setting \citep{garivier2008upper, koulouriotis2008reinforcement, besbes2014stochastic}. The non-stationary rewards for the behavior, that decay over time, have been considered for the bandit setting, under rotting bandits \citep{levine2017rotting,seznec2019rotting}; these algorithms do not obviously extend to the RL setting. 

\section{Handling the Non-Stationarity in a Multi-prediction System}

In this section, we describe a unified approach to handle non-stationarity in both the GVF and behavior learners, using successor features. We first discuss how to use successor features to learn under non-stationary cumulants, for prediction. Then we discuss using successor features for control, allows us to leverage this approach for non-stationary rewards for the behavior. We then discuss state-reweightings, and how to mitigate non-stationarity due to a changing behavior. 

\subsection{Successor Features for Non-stationary Rewards}\label{sec:nr-sf}

% Assume we are given the successor features, and describe that we only have to track rewards. Let's call this Successor Features for Non-stationary Rewards (SF-NR). Point to later section where we prove this is more effective

Successor features provide an elegant way to learn value functions under non-stationarity. The separation of learning stationary successor features and rewards enables more effective tracking of non-stationary rewards, as we explain in this section and formally prove in Section \ref{sec:theory}.

Assume that there is a weight vector $\wvec^* \in \RR^\numfeats$ and features $\xvec(s,a) \in \RR^\numfeats$ for each state and action $(s,a)$ such that $r(s,a) = \langle \xvec(s,a), \wvec^*\rangle$. Recursively define
\begin{align*}
\psivec(s,a) = \mathbb{E}_\pi[\xvec(S_t, A_t) 
+ \gamma_{t+1} \psivec(S_{t+1}, A_{t+1}) | S_t =s, A_t =a]
\end{align*}
$\psivec(s,a)$ is called the \emph{successor features}, the discounted cumulative sum of feature vectors, if we follow policy $\pi$. For $\psivec_t \defeq \psivec(S_t,A_t)$ and $\xvec_t \defeq \xvec(S_t,A_t)$, we can see $Q(s,a) = \langle \psivec(s,a), \wvec^*\rangle$ 
\begin{align*}
&\langle \psivec(s,a), \wvec^*\rangle 
= \mathbb{E}_\pi[\langle \xvec_t, \wvec^*\rangle | S_t =s, A_t =a] + \mathbb{E}_\pi[\gamma_{t+1} \langle \psivec_{t+1}, \wvec^* \rangle | S_t =s, A_t =a]\\
&= r(s, a) + \mathbb{E}_\pi[\gamma_{t+1} \langle \xvec_{t+1}, \wvec^*\rangle | S_t =s, A_t =a]+ \mathbb{E}_\pi[\gamma_{t+1} \gamma_{t+2} \langle \psivec_{t+2}, \wvec^* \rangle | S_t =s, A_t =a]\\
&= r(s, a) + \mathbb{E}_\pi[\gamma_{t+1} r_{t+1} | S_t =s, A_t =a] + \mathbb{E}_\pi[\gamma_{t+1} \gamma_{t+2} \langle \psivec_{t+2}, \wvec^* \rangle | S_t =s, A_t =a]\\
&= \quad \ldots
\quad = \mathbb{E}_\pi[r(s, a) + \gamma_{t+1} r_{t+1} + \gamma_{t+1} \gamma_{t+2} r_{t+2} + \ldots | S_t =s, A_t =a]\quad = Q(s,a).
\end{align*}
% MARTHAC: Left old double column format in the case of inevitable rejection
%\begin{align*}
%&\langle \psivec(s,a), \wvec^*\rangle 
%= \mathbb{E}_\pi[\langle \xvec(S_t, A_t), \wvec^*\rangle | S_t =s, A_t =a] \\
%&\ \ \ \ \ + \mathbb{E}_\pi[\gamma_{t+1} \langle \psivec(S_{t+1}, A_{t+1}), \wvec^* \rangle | S_t =s, A_t =a]\\
%&= r(s, a) + \mathbb{E}_\pi[\gamma_{t+1} \langle \xvec(S_{t+1}, A_{t+1}), \wvec^*\rangle | S_t =s, A_t =a]\\
%&\ \ \ \ \ + \mathbb{E}_\pi[\gamma_{t+1} \gamma_{t+2} \langle \psivec(S_{t+2}, A_{t+2}), \wvec^* \rangle | S_t =s, A_t =a]\\
%&= r(s, a) + \mathbb{E}_\pi[\gamma_{t+1} r(S_{t+1}, A_{t+1}) | S_t =s, A_t =a]\\
%&\ \ \ \ \  + \mathbb{E}_\pi[\gamma_{t+1} \gamma_{t+2} \langle \psivec(S_{t+2}, A_{t+2}), \wvec^* \rangle | S_t =s, A_t =a]\\
%&= \ldots\\
%&= \mathbb{E}_\pi[r(s, a) + \gamma_{t+1} r(S_{t+1}, A_{t+1})\\
%& \ \ \ \ \ + \gamma_{t+1} \gamma_{t+2} r(S_{t+2}, A_{t+2}) + \ldots | S_t =s, A_t =a]\\
%&= Q(s,a)
%\end{align*}
%
If we have features $\xvec(s,a) \in \RR^d$ which allow us to represent the immediate reward, then successor features provide a good representation to approximate the GVF. 
%To use successor features to estimate the value function, 
We simply learn another set of parameters $\cparams \in \RR^\numfeats$ that predict the immediate cumulant (or reward): $\cfunc(s,a) \approx \langle \xvec(s,a), \cparams \rangle$. These parameters $\cparams$ are updated using a standard regression update, and $Q(s,a) \approx \langle \psivec(s,a), \cparams\rangle$ .

The successor features $\psivec(s,a)$ themselves, however, also need to be approximated. 
In most cases, we cannot explicitly maintain a separate $\psivec(s,a)$ for each $(s,a)$, outside of the tabular setting.
% and instead must approximate $\psivec(s,a)$. 
Notice that each element in $\psivec(s,a)$ corresponds to a true expected return: the cumulative discounted sum of a reward feature into the future.
% MARTHAC: I wrote this, but admittedly it is repetitive
%\footnote{Notice that $\psivec(s,a)$ is called the successor features because it encodes the \emph{reward} features into the future. But we do not use the vector $\psivec(s,a)$ explicitly as a set of features to learn the values, because $\psivec(s,a)$ is an idealized quantity that we approximate and because we would not use these features directly to estimate the value with temporal difference algorithms. Instead, we estimate the weights to learn the immediate rewards.} 
Therefore, $\psivec(s,a)$ can be approximated using any value function approximation method, such as temporal difference (TD) learning.
% using function approximation to output an estimate of the vector $\psivec(s,a)$ for a given $(s,a)$. 
%as each entry corresponds to a cumulative discounted sum.
We learn parameters $\psiparams$ for the approximation $\hat\psivec(s,a; \psiparams) = [\hat\psivec_1(s,a; \psiparams), ..., \hat\psivec_d(s,a; \psiparams)]^\top \in \RR^d$ where $\hat\psivec_m(s,a; \psiparams) \approx \psivec_m(s,a)$. We can use any function approximator for $\hat\psivec(s,a; \psiparams)$, such as linear function approximation with tile coding with $\psiparams$, linearly weighting the tile coding features to produce $\hat\psivec(s,a; \psiparams)$, or neural networks, where $\psiparams$ are the parameters of the neural network. %The parameters $\psiparams$ can be updated using any GVF learning algorithm. 

We summarize the algorithm using successor features for non-stationary rewards/cumulants, called SF-NR, in Algorithm \ref{alg:sf-nr}.
We provide an update formula for the approximate SF using Expected Sarsa for prediction \citep{suttonbartobook} for simplicity, but note that any value learning algorithm can be used here. In our experiments, we use Tree-Backup \citep{precup2000eligibility} because it reduces variance from off-policy learning; we provide the pseudocode in Appendix~\ref{app:algs}. Algorithm \ref{alg:sf-nr} assumes that the reward features $\xvec(s,a)$ are given, but of course these can be learned as well. Ideally, we would learn a compact set of reward features that provide accurate estimates as a linear function of these reward features. A compact (smaller) set of reward features is preferred because it makes the SF more computationally efficient to learn.

\begin{wrapfigure}[13]{r}{0.43\textwidth}
\vspace{-1.3cm}
\begin{minipage}{0.43\textwidth}
 \begin{algorithm}[H]
   \caption{Successor Features for \\Non-stationary Rewards (SF-NR)}
   \label{alg:sf-nr}
   \! {\bfseries Input:} \!\!$(S_t, \!A_t, \!S_{t+1}, \!C_{t+1}, \!\gamma_{t+1})$, \!$\pi$, \!$\psiparams$, \!$\cparams$
 \begin{algorithmic}
  \STATE $\xvec \gets \xvec(S_t,A_t)$
  \STATE $\hat\psivec \gets \hat\psivec(S_t,A_t;\psiparams)$
%   \STATE $\zvec \gets \gamma_t \pi(A_t|S_t) \lambda \zvec_{t-1} + \xvec$ 
  \STATE $\hat\psivec' \gets \sum_{a'} \pi(a'|S_{t+1}) \hat\psivec(S_{t+1},a';\psiparams)$
  \STATE $\Delta \gets \vec{0}$
   \FOR{$m=1$ {\bfseries to} $\numfeats$}
	\STATE $\delta_m \gets \xvec_m + \gamma_{t+1} \hat\psivec'_m - \hat\psivec_m$
	\STATE $\Delta \gets \Delta + \delta_m \nabla \hat\psivec_m$
   \ENDFOR	
   \STATE $\psiparams \gets \psiparams + \alpha \Delta$
   \STATE $\cparams \gets \cparams + \alpha (C_{t+1} - \langle \xvec, \cparams \rangle) \xvec$
 \end{algorithmic}
 \end{algorithm}
\end{minipage}
\end{wrapfigure}
There are two key advantages from the separation into learning successor features and immediate cumulant estimates. First, it easily allows different or changing cumulants to be used, for the same policy, using the same successor features. The transition dynamics summarized in the stationary successor features can be learned slowly to high accuracy and re-used. 
%If we want to learn the action-values for a new cumulant, then we only have to solve a simpler regression problem of predicting the immediate cumulant. The transition dynamics summarized in the successor features can be re-used. 
This re-use property is why these representations have been used for transfer \citep{barreto2017successor,barreto2018transfer,barreto2020fast}. This property is pertinent for us, because it allows us to more easily track changes in the cumulant. The regression updates can quickly update the parameters $\cparams$, and exploit the already learned successor features to more quickly track value estimates. Small changes in the rewards can result in large changes in the values; without the separation, therefore, it can be more difficult to directly track the value estimates. 

Second, the separation allows us to take advantage of online regression algorithms with strong convergence guarantees. Many optimizers and accelerations are designed for a supervised setting, rather than for temporal difference algorithms. Once the successor features are learned, the prediction problem reduces to a supervised learning problem. We can therefore even further improve tracking by leveraging these algorithms to learn and track the immediate cumulant. We formalize the convergence rate improvements, from this separation, in Section \ref{sec:theory}. 
%The overall algorithm, which we call Successor Features for Nonstationary Reward (SF-NR), is summarized in Algorithm \ref{alg:sf-nr}.

%For each GVF, we maintain successor feature weights $\psiparams$ and cumulant weights $w_c$. The successor feature weights will converge, as long as the state distribution weighting $d$ is not changing. The cumulant weights $w_c$ are constantly updated towards cumulant targets. If the cumulants are stationary, then we expect these weights to stop changing at some point. If the cumulants are drifting, then we expect the weights $w_c$ to constantly change. The value function estimate, however, can be take advantage of the learned successor features, which do not change.  We formalize this faster learning in Section \ref{sec:theory}.

\subsection{GPI with Successor Features for Control}

In this section we outline a control algorithm under non-stationary rewards. SF-NR provides a method for updating the value estimate due to changing rewards. The behavior for the multi-prediction problem has changing rewards, and so could benefit from SF-NR. But SF-NR only provides a mechanism to efficiently track action-values for a fixed policy, not for a changing policy. Instead, we turn to the idea of constraining the behavior to act greedily with respect to the values for a set of policies, introduced as Generalized Policy Improvement (GPI) \cite{barreto2018transfer,barreto2020fast}. 
%which we call policy-constrained greedification (PCG), where the behavior is greedy only over these policies rather than actions. 

For our system, this is particularly natural, as we are already learning successor features for a collection of policies. Let us start there, where we assume our set of policies is $\Pi = \{\pi_1, \ldots, \pi_\numpred\}$. Assume also that we have learned the successor features for these policies, $\hat\psivec(s,a; \psiparams^{(\predind)})$, and that we have weights $\bwr \in \RR^\numfeats$ such that $\langle \xvec(s,a), \bwr \rangle \approx \mathbb{E}[R_{t+1} | S_t = s, A_t = a]$ for behavior reward $R_{t+1}$. 
%The action-value estimate for these rewards, under policy $\pi^{(\predind)}$ is $\hat Q^{(\predind)}_r(s,a) \defeq \langle \hat\psivec(s,a; \psiparams^{(\predind)}), \bwr \rangle$. 
Then on each step, the behavior policy takes the following greedy action
\begin{align*}
\bpolicy(s) =  \argmax_{a} \max_{\predind \in \{1, \ldots, \numpred\}} \hat Q^{(\predind)}_r(s,a) = \argmax_{a} \max_{\predind \in \{1, \ldots, \numpred\}} \langle \hat\psivec(s,a; \psiparams^{(\predind)}), \bwr \rangle
\end{align*}    
The resulting policy is guaranteed to be an improvement: in every state the new policy has a value at least as good as any of the policies in the set \citep[Theorem 1]{barreto2017successor}. Later work also showed sampled efficiency of GPI when combining known reward weights to solve novel tasks \citep{barreto2020fast}.

The use of successor features has similar benefits as discussed above, because the estimates can adapt more rapidly as the rewards change, due to learning progress changing over time. The separation is even more critical here, as we know the rewards are constantly drifting, and tracking quickly is even more critical. We could even more aggressively adapt to these non-stationary rewards, by anticipating trends. For example, instead of a regression update, we can model the trend (up or down) in the reward for a state and action. If the reward has been decreasing over time, then likely it will continue to decrease. Stochastic gradient descent will put more weight on recent points, but would likely predict a higher expected reward than is actually observed. For simplicity here, we still choose to use stochastic gradient descent, as it is a reasonably effective tracking algorithm, but note that performance improvements could likely be obtained by exploiting this structure in this problem. 

We can consider a different set of policies for GVFs and behavior. However, the two are naturally coupled. 
%Consider the case where the behavior uses a different set of policies for PCG. 
First, the GPI theory shows that greedifying over a larger collection of policies provides better policies. It is sensible then to at least include the GVF policies into the set for the behavior. Second, the behavior needs to learn the successor features for the additional policies. Arguably, it should try to gather data to learn these well, so as to facilitate its own policy improvement. It should therefore also incorporate the learning progress for these successor features, into the intrinsic reward. For this work, therefore, we assume that the behavior uses the set of GVF policies. Note that the weight change intrinsic reward uses the concatenation of $\psiparams$ and $\cparams$.
%We therefore assume that the two sets of policies are the same.\footnote{Note that the intrinsic reward is defined as the weight change. For GVF learners that use SF-NR, the weights include both SR weights and cumulant weights $\tweights = [\psiparams, \cparams]$. For policies where the only goal is to learn the SR, then the change in weights would just be for $\psiparams$.}

\subsection{Interest and prior corrections for the changing state distribution} \label{sec:prior_corrections}

%The previous two sections described the general strategies for the prediction learners and the behavior learner. We provided a specific TD algorithm, to be concrete, but highlighted than any TD variant could be used. In this section, we argue that to obtain stationary successor features, we need to use an algorithm with prior corrections. The key issue is that, without prior corrections, the state-weighting changes as the behavior changes, meaning the objective is non-stationary. 

The final source of non-stationarity is in the state distribution. As the behavior $\bpolicy$ changes, the state-action visitation distribution $\bdistt: \States \times \Actions \rightarrow [0,1]$ changes. 
% MARTHAC: What?
%When updating online, 
The state distribution implicitly weights the relative importance of states in the GVF objective, called the projected Bellman error (PBE). Correspondingly, the optimal SF solution could be changing, since the objective is changing. The impact of a changing state-weighting depends on function approximation capacity, because the weighting indicates how to trade-off function approximation error across states. When approximation error is low or zero---such as in the tabular setting---the weighting has little impact on the solution. Generally, however, we expect some approximation error and so a non-negligible impact. 
%For example, in a tabular setting, the agent can achieve zero BE in every state. In general, however, we will have approximation error and so should consider this source of error. 

% MARTHAC: Maybe we have no space for this
%The impact of this changing state distribution depends on the function approximation capacity. The weighting indicates how to trade-off function approximation error across states. When approximation error is low or zero, the weighting has little impact on the solution. For example, in a tabular setting, the agent can achieve zero BE in every state. In general, however, we will have approximation error and so should consider this source of error. 

We can completely remove this source of non-stationary by using \emph{prior corrections}. These are products of importance sampling ratios, that reweight the trajectory to match the probability of seeing that trajectory under the target policy $\pi$. Namely, it modifies the state weighting to $d_\pi$, the state-action visitation distribution under $\pi$. We explicitly show this in Appendix~\ref{app:prior-correction-pbe}. 
Unfortunately, prior corrections can be highly problematic in a system where the behavior policy takes exploratory actions and target policies are nearly deterministic. It is likely that these corrections will often either be zero, or near zero, resulting in almost no learning. 

To overcome this inherent difficulty, we restrict which states are important for each predictive question. Likely, when creating a GVF, the agent is interested in predictions for that GVF only in certain parts of the space. This is similar to the idea of initiation sets for options, where an option is only executed from a small set of relevant states. We can ask: what is the GVF answer, from this smaller set of states of interest? This can be encoded with a non-negative interest function, $i(s,a)$, where some (or even many) states have an interest of zero. This interest is incorporated into the state-weighting in the objective, so the agent can focus function approximation resources on states of interest. 

When using interest, it is sensible to use emphatic weights \citep{sutton2016emphatic}. Emphatic weightings are a prior correction method, used under the excursions model \citep{patterson2021generalized}. They reweight to a discounted state-action visitation under $\pi$ when starting from states proportionally to $\bdist$. Further, they ensure states inherit the interest of any states that bootstrap off of them. Even if a state has an interest of zero, we want to accurately estimate its value if an important states bootstraps off of its value. The combination of interest and emphatic weightings---which shift state-action weighting to visitation under $\pi$---means that we mitigate much of the non-stationarity in the state-action weighting. We provide the pseudocode for this Emphatic TB (ETB) algorithm in Appendix~\ref{app:algs}.

%For us, it helps mitigate the issue of a changing state weighting, because $d_{b_t}(s,a)$ is zeroed by $i(s,a)$ in many states. states with zero-interest need not have function approximation resources wasted on   Restricting GVFs with 
%A practical alternative is to use interest functions that restrict state weighting and emphatic weights \citep{sutton2016emphatic}. Emphatic weightings are a prior correction method, used under the excursions model \citep{patterson2021generalized}. They reweight to a discounted state-action visitation under $\pi$ when starting from states proportionally to $\bdist$. They also facilitate using an interest function $i(s,a)$, allowing some state weightings to be zero. The combination of interest and shifting state-action weighting to visitation under $\pi$ means that we mitigate much of the non-stationarity in the state-action weighting. We provide the pseudocode for this Emphatic TB (ETB) algorithm in Appendix~\ref{app:algs}.

% MARTHA: Are any of these citations useful?
%Off-policy temporal difference prediction algorithms have been widely studied \cite{lagoudakis2003least, suttonbartobook, sutton1988learning, precup2000eligibility, ghiassian2020gradient, geist2014off}. In this work, we evaluate off-policy Expected Sarsa($\lambda$) (ES($\lambda$)) \cite{suttonbartobook} and Tree Backup($\lambda$) (TB($\lambda$)) \cite{precup2000eligibility}. We give a brief description of both algorithms below, but a more detailed description can be found in \textit{Reinforcement Learning: An Introduction} \cite{suttonbartobook}.

\section{Sample Efficiency of SF-NR }
\label{sec:theory}

As suggested in Section~\ref{sec:nr-sf}, the use of successor features
makes SF-NR particularly well-suited
to our multi-prediction problem setting.
The reason for this is simple: given
access to an accurate SF matrix, value function estimation reduces
to a \textit{fundamentally simpler} linear prediction problem. Indeed,
access to an accurate SF enables one to
sidestep known lower-bounds on PBE estimation. 
%This is particularly important when non-stationarity of one-or-more of the reward functions necessitates periodically revising the value estimates.

For simplicity, we prove the result for value functions; the result easily extends to action-values. Denote by $\vvec^{\pi}\in\R^{\Abs{\cS}}$ the vector with
entries $v^{\pi}(s)$, $\rvec ^{\pi}\in\R^{\Abs{\cS}}$ the vector of expected immediate rewards in each state,
and $\Pmat\in\R^{\Abs{\cS}\times\Abs{\cS}}$ the matrix of transition probabilities.
%Given a state weighting $d:\cS\to\R$, let $\Dmat=\text{Diag}(\Set{d(s)}_{s\in\cS})$ and let $\norm{\cdot}_{\Dmat}$ denote the weighted norm under $\Dmat$. 
The following lemma, proven in Appendix~\ref{app:msve-bound}, relates  mean squared value error (VE) to one-step reward prediction error.
\vspace{-0.3cm}
\begin{restatable}{lemma}{MSVEBound}\label{lemma:msve-bound}
  Assume there exists a $\wvec^{*}\in\R^{d}$ such that $\rvec^{\pi}=\Xmat \wvec^{*}$.
  Let $\hat \rvec\defeq\Xmat\wvec$ for some $\wvec\in\R^{d}$, and
  let $\Dmat=\text{Diag}(\{d(s)\}_{s\in\cS})$ for distribution $d$ fully supported on $\cS$, with $\norm{\cdot}_{\Dmat}$ the weighted norm under $\Dmat$.
  Then the value estimate $\hat \vvec\defeq\Psi\wvec$ satisfies
  $\half\norm{\vvec^{\pi}-\hat\vvec}^{2}_{\Dmat}\le\frac{\norm{\rvec^{\pi}-\hat\rvec}_{\Dmat}^{2}}{2(1-\gamma)^{2}}$.
%  \begin{align}
%    \half\norm{\vvec^{\pi}-\hat\vvec}^{2}_{\Dmat}\le\frac{\norm{\rvec^{\pi}-\hat\rvec}_{\Dmat}^{2}}{2(1-\gamma)^{2}}\label{eq:vrate}
%    % \E_{s\sim d}\sbrac{\half\brac{v^{\pi}(s)-\hat v(s)}^{2}}\le\frac{\E_{s\sim d}\sbrac{\half(r(s)-\hat r(s))^{2}}}{(1-\gamma)^{2}}
%  \end{align}
\end{restatable}
\vspace{-0.1cm}
%
%Assuming the rewards are realizable as a function of the features,  $r(s) = \inner{\xvec(s)}{\wvec^{*}}$ for
%some $\wvec^{*}\in\R^d$,
Thus we can ensure that $\MSVE(\vvec^{\pi},\hat \vvec)\le \eps$ by ensuring that
$\norm{\rvec^{\pi}-\hat\rvec}_{\Dmat}^{2}\le\eps(1-\gamma)^{2}$. This is promising,
as this latter expression is readily expressed as the objective of a linear
regression problem.
%\begin{align}
%  \cL(\wvec)\defeq\half\norm{\rvec^{\pi}-\hat\rvec}_{D}^{2} = \E_{s\sim d}\sbrac{\half(r(s)-\inner{\xvec(s)}{\wvec})^{2}}\label{eq:sqrloss}.
%\end{align}
To illustrate the utility of this, let's look at a concrete example: suppose the agent has an accurate SF matrix $\Psi$ and
that the reward function changes at some point in the agent's deployment. 
%necessitating correction of our value estimates. 
Suppose access to a batch of transitions $\cD\defeq\Set{S_{t},A_{t},S_{t}^{\prime},r_{t},\rho_{t}}_{t=1}^{T}$ with which we can correct our
estimate of $v^{\pi}$, where each $(s,a,s^{\prime},\rho)\in\cD$ is such that
$s\sim \bdist$ for some known behavior policy $\bpolicy$, $A_{t}\sim\pi(\cdot|s)$,
$s^{\prime}\sim P(\cdot|s,a)$ and $r = r(s,a,s^{\prime})$.
% Assume that the samples are collected such that $S_{t}\sim d_{\mu}$ for some behavior policy $\mu$, $A_{t}\sim\pi(\cdot|S_{t})$,
% $S_{t}^{\prime}\sim P_{\pi}(\cdot | S_{t}, A_{t})$, $r_{t}=r(S_{t},A_{t},S_{t}^{\prime})$, and $\rho_{t}=\frac{\pi(A_{t}|S_{t})}{\mu(A_{t}|S_{t})}$.
Assume for simplicity that $\rho_{t}\le\rho_{\max{}}$,
$\norm{\phi(S_{t})}_{\infty}\le L$, $r_{t}\le R_{\max{}}$ for some finite $\rho_{\max{}},R_{\max{}},L\in\R_{+}$.
%If the agent has access to an accurate SF matrix $\Psi$, we can ensure $\hat v\rightarrow v^{\pi}$ by minimizing Equation \ref{eq:sqrloss} w.r.t parameters $w$ and using value estimate $\hat v(s)=\inner{\psi(s)}{w}$.
Then we can get the following result, proven in Appendix~\ref{app:regret-bound}, that is a straightforward application of \citet[Theorem 7.26]{orabona2019modern}. 
\vspace{-0.3cm}
\begin{restatable}{proposition}{RegretBound}\label{prop:regret-bound}
Define $\ell_{t}(w)\defeq \frac{\rho_{t}}{2}\brac{r_{t}-\inner{x(S_{t})}{w}}^{2}$. Suppose we apply a basic recursive least-squares estimator to minimize regret on this loss sequence, producing a sequence of iterates $w_{t}$. Let $\wbar_{T}\defeq\frac{1}{T}\sum_{t=1}^{T}w_{t}$ denote the average iterate. For $\hat v(s) = \inner{\psi(s)}{\wbar_{T}}$, we have that
\begin{align}
  \norm{\vvec^{\pi}-\hat\vvec}^{2}_{\Dmat}
  %&\qquad\le \frac{\norm{w^{*}}^{2}+dR_{\max{}}^{2}\Log{1+\frac{\rho_{\max{}}L^{2}T}{d}}}{2(1-\gamma)^{2}T}\\
  \le O\brac{\frac{d \rho_{\max}R_{\max{}}^{2}\Log{1+\rho_{\max{}}L^{2}T}}{(1-\gamma)^{2}T}} \label{eq:sf-bound}.
\end{align}
\end{restatable}
\vspace{-0.3cm}
In contrast, without the SF we are faced with minimizing a harder objective: the PBE.
% In contrast, without the SF we're faced with a problem which involves bootstrapping, off-policy learning, and function approximation --- a setting in which
% regular TD methods are known to diverge \citep{baird1995residual}.
It can be shown that minimizing the PBE is equivalent to a stochastic saddle-point problem, and
the convergence to the saddle-point of this problem has an unimprovable rate of $O\brac{\frac{\tau}{T^{2}}+\frac{(1+\gamma)\rho_{\max{}}L^{2}d}{T} + \frac{\sigma}{\sqrt{T}}}$ where $\tau$ is the maximum eigenvalue of the covariance matrix and $\sigma$ bounds gradient stochasticity, %$\tau=\lambda_{\max}(\mathbf{C})$ 
and this convergence rate translates into the performance bound 
$\half\norm{\vvec^{\pi}-\hat \vvec}^{2}_{\Dmat}\le O\brac{\sqrt{\frac{\tau}{T^{2}}+\frac{(1+\gamma)\rho_{\max{}}L^{2}d}{T}+\frac{\sigma}{\sqrt{T}}}}$  \citep[Proposition 5]{liu2018proximal}.
%\begin{align}
%  \half\norm{\vvec^{\pi}-\hat \vvec}^{2}_{\Dmat}\le O\brac{\sqrt{\frac{\tau}{T^{2}}+\frac{(1+\gamma)\rho_{\max{}}L^{2}d}{T}+\frac{\sigma}{\sqrt{T}}}} \label{eq:gtdbound}
%\end{align}
%\footnote{
  %To see this, substitute the optimal rate of convergence in the saddle-point problem into Proposition 5 of \citet{liu2018proximal}.
%}.
Comparing with Equation \ref{eq:sf-bound}, we observe an additional dependence
of $O(\sqrt{\tau}/T)$ as well as the worse dependence of at least
$O(1/\sqrt{T})\ge(\Log{T}/T)$ on all other quantities of interest, reinforcing
the intuition that access to the SF enables us to more efficiently re-evaluate
the value function.

\section{A First Experiment Testing the Multi-prediction System}

\begin{wrapfigure}[14]{l}{0.44\textwidth}
\vspace{-0.3cm}
  \begin{centering}
      \includegraphics[width=0.4\textwidth]{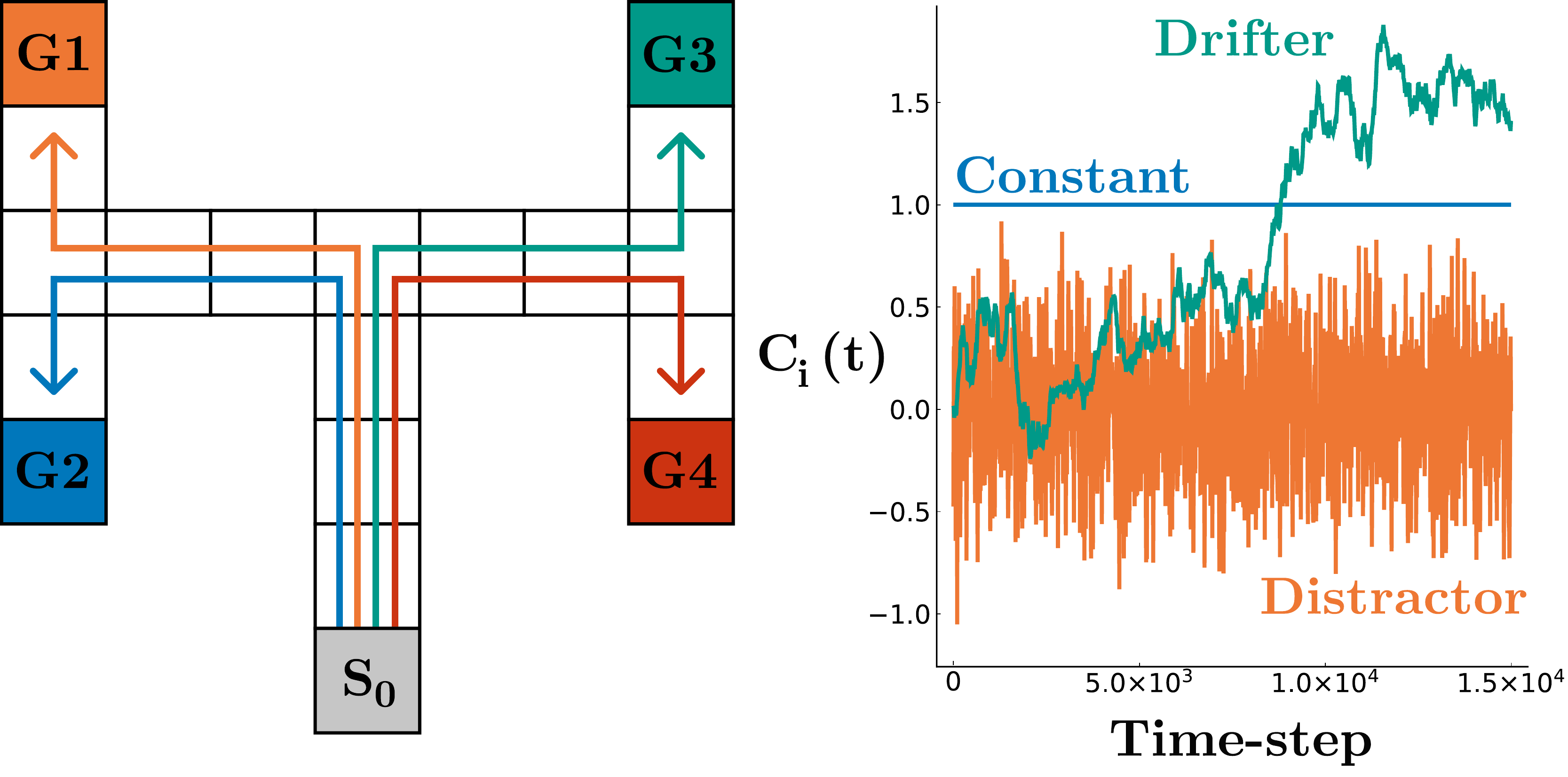}
  \end{centering}
  \vspace{-0.3cm}
    \caption{\label{fig:objectives}
      \textbf{Tabular TMaze} with 4 GVFs, with cumulants of zero except in the goals. The right plot shows the cumulants in the goals. G2 and G4 have constant cumulants, G1 has a distractor cumulant and G4 a drifter. 
    }\label{fig_tmaze}  
\end{wrapfigure}
In this section, we investigate the utility of using SF-NR under non-stationary cumulants and rewards, both for prediction and control. We conduct the experiment in a TMaze environment, inspired by the environments used to test animal cognition \citep{tolman1930introduction}. The environment, depicted in Figure \ref{fig_tmaze}, has four GVFs where each policy takes the fastest route to its corresponding goal. The cumulants are zero everywhere except for at the goals. The cumulant can be of three types: a constant fixed value (\textbf{constant}), a fixed-mean and high variance value (\textbf{distractor}), or a non-stationary zero-mean random walk process with a low variance (\textbf{drifter}). Exact formulas for these cumulants are in Appendix~\ref{app:tmazedetails}.

%We conducted several experiments to evaluate the efficiency of our algorithm in a multi-prediction problem called the \emph{TMaze}. In the TMaze, each GVF corresponds to a goal state which provides a possibly non-stationary cumulant that the agent receives once its arrived, and the target policy is the direct path towards that goal state. The primary feature of the TMaze environment is a tree-like structure with junctions that split off many times until the end of a hallway is reached, where a GVF's goal state is located. This structure enables the generated experience to be relevant in learning multiple prediction problems.

\textbf{Utility of SF-NR for a Fixed Behavior Policy}

We start by testing the utility of SF-NR for GVF learning, under a fixed policy that provides good data coverage for every GVF. The \emph{Fixed-Behavior Policy} is started from random states in the TMaze, and moves towards the closest goal, with a 50/50 chance of going either direction if there is a tie. This policy is like a round robin policy, in that one of the GVF policies is executed each episode and, in expectation, all four policies are executed the same number of times. 

We compare an agent that uses SF-NR and one that learns the approximate GVFs using Tree Back-Up (TB). TB is an off-policy temporal difference (TD) algorithm, that reduces variance in the eligibility trace. We also use TB to learn the successor features in SF-NR. Both use $\lambda = 0.9$ and a stepsize method called Auto \citep{mahmood2012tuning} designed for online learning. We sweep the initial stepsize and meta stepsizes for Auto. For further details about the agents and optimizer, see Appendix~\ref{app:algs}. We additionally compare to least squares TD (LSTD), with $\lambda = 0.9$, particularly as it computes a matrix similar to the SF, but does not separate out cumulant learning (see Appendix~\ref{app:relationship-srnf-td} for this connection). 

\begin{figure}[t]
 	\begin{subfigure}[b]{0.31\textwidth}
         \centering
         \includegraphics[width=\textwidth]{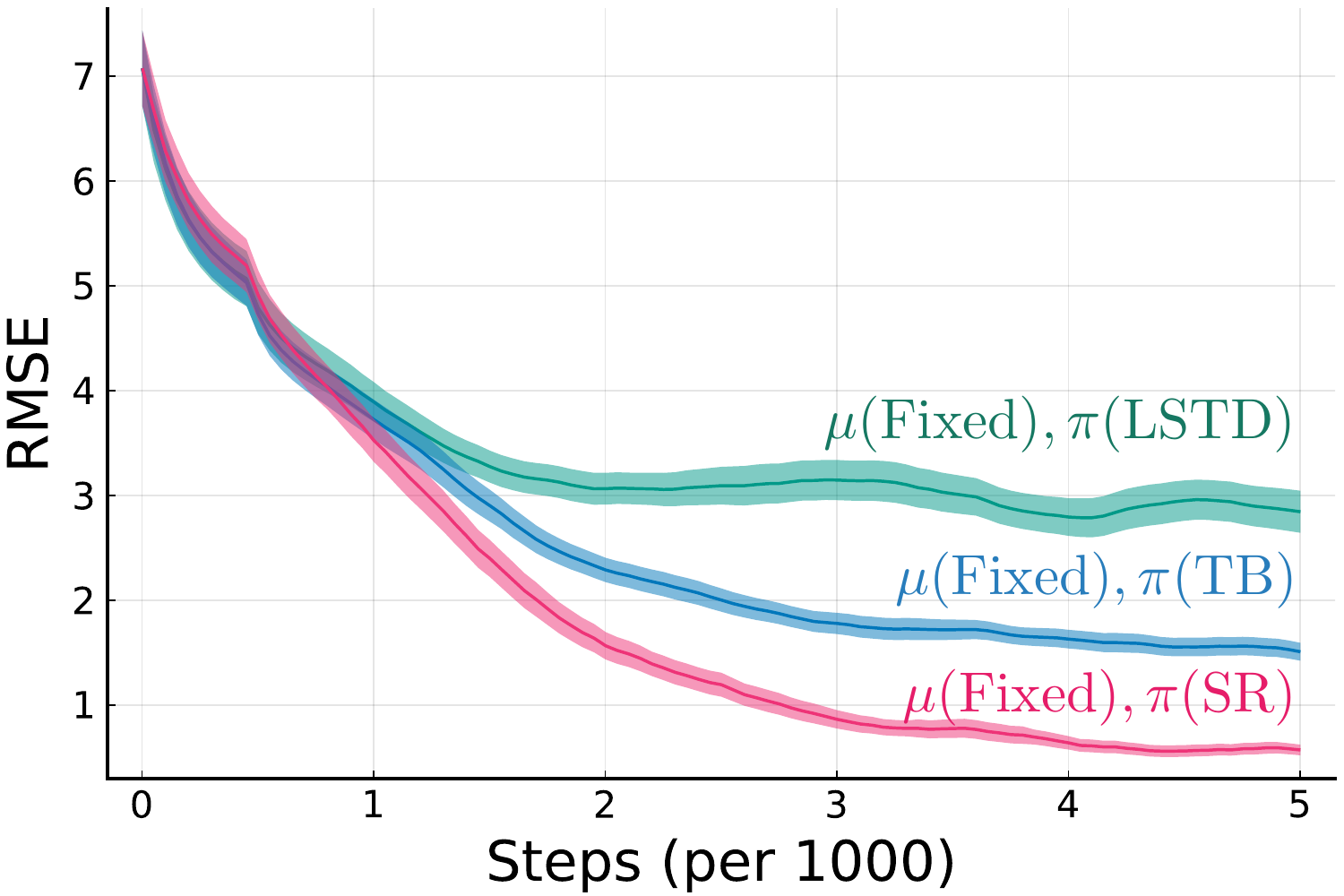}
         \caption{Fixed-Behavior Policy}
         \label{fig:round_robin_tmaze_rmse}
     \end{subfigure}
 	\begin{subfigure}[b]{0.31\textwidth}
         \centering
         \includegraphics[width=\textwidth]{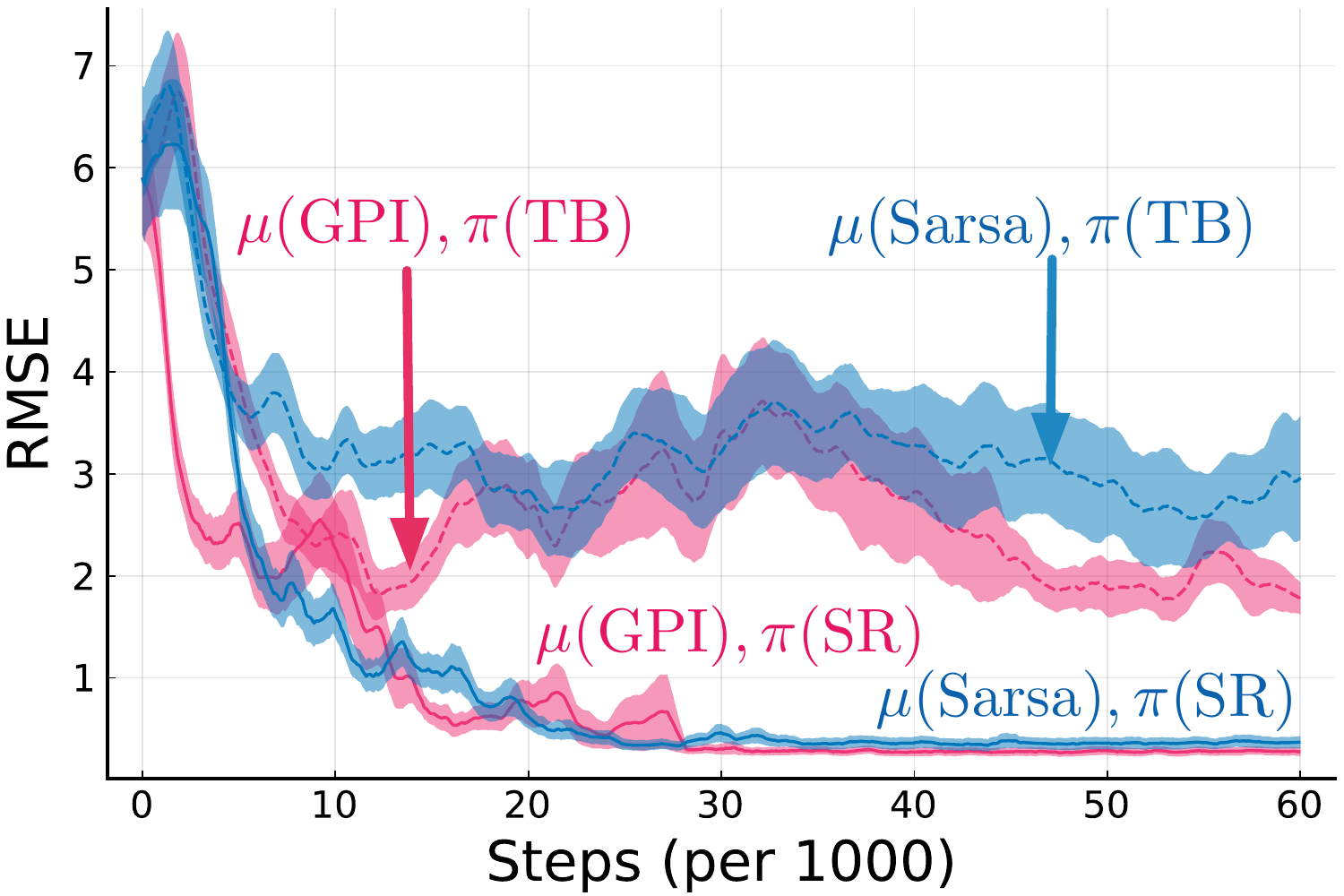}
         \caption{Learned Behavior Policy}
         \label{fig:control_tmaze_rmse}
     \end{subfigure}     
 	\begin{subfigure}[b]{0.37\textwidth}
         \centering
         \includegraphics[width=0.9\textwidth]{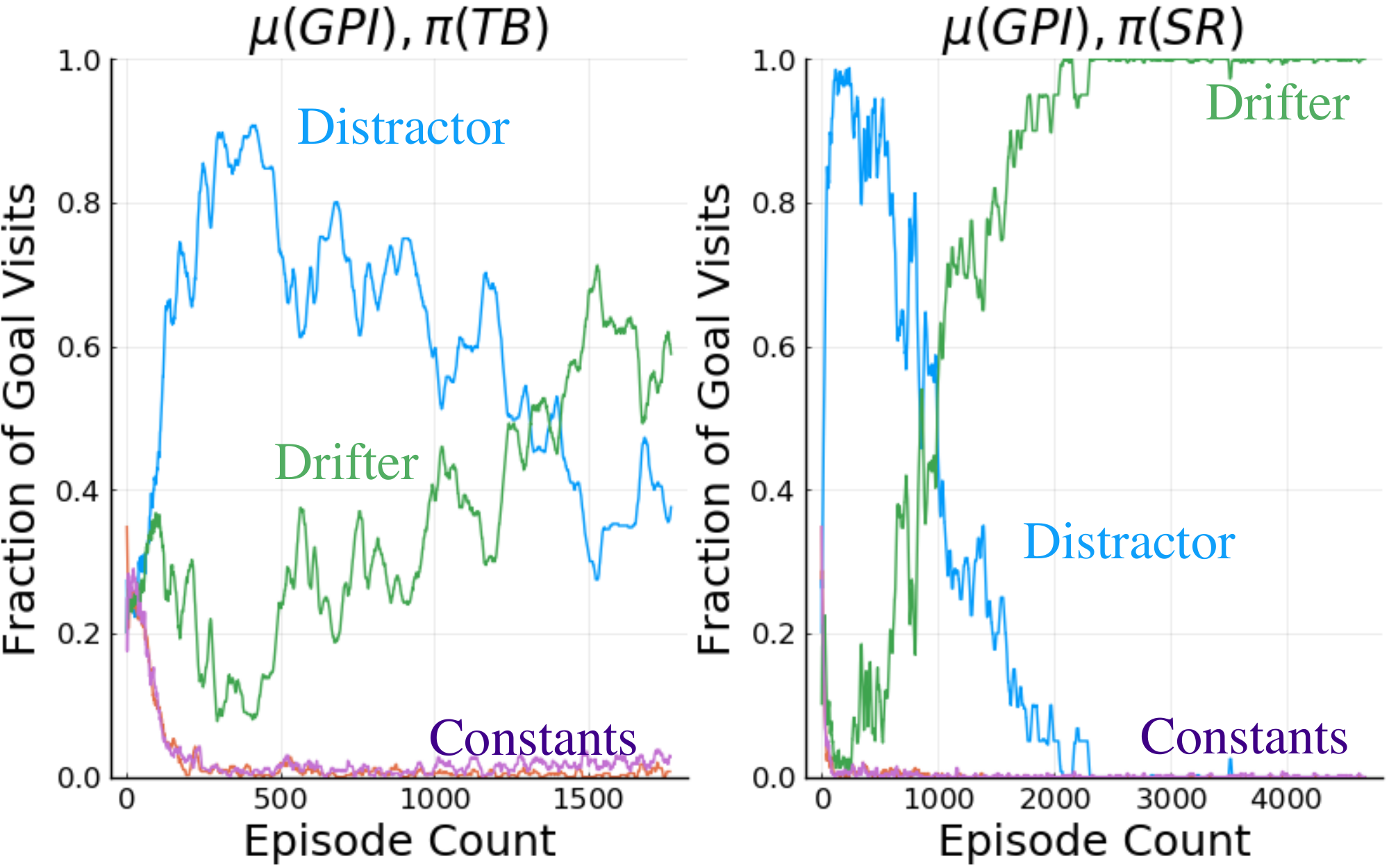}
         \caption{Visitation Plot Comparison}
         \label{fig:visitation_tmaze}
     \end{subfigure}      
     \caption{Performance in \textbf{Tabular TMaze}, with averages over 30 runs. \textbf{(a)} and \textbf{(b)} show average off-policy prediction RMSE, with standard errors, where the error is weighted by (a) the state distribution $\bdist$ for the Fixed-Behavior policy and (b) a uniform state weighting when learning the behavior. 
     \textbf{(c)} Goal visitation plots for GPI with SF and TB.}
     \vspace{-0.4cm}
\end{figure}

In Figure \ref{fig:round_robin_tmaze_rmse}, we can see SF-NR allows for much more effective learning, particularly later in learning when it more effectively tracks the non-stationary signals. LSTD performs much more poorly, likely because it corresponds to a closed-form batch solution, which uses old cumulants that are no longer reflective of the current cumulant distribution.  
% compared to their Q learning counterparts. The SF-NR demons and $TB(\lambda)$ demons are using Auto for adaptive step size as this results in lower RMSE and is central for introspective learners. See Appendix for analysis of adaptive vs non-adaptive step size methods.

\textbf{Investigating GPI for Learning the Behavior}

Next we investigate if SF-NR improves learning of the whole system, both for the GVFs and for the behavior policy. We use SF-NR and TB for the GVF learners, and Expected Sarsa (Sarsa) and GPI for the behavior. The GPI agent uses the GVF policies for its set of policies. The reward features for the behavior are likely different than those for the GVF learners, because the cumulants are zero in most states whereas intrinsic rewards are likely non-zero in most states. The GPI agent, therefore, learns its own SFs for each policy, also using TB. The reward weights that estimate the (changing) intrinsic rewards are learned using Auto, as are the SFs. Note that the behavior and GVF learners all share the same meta-step size---namely only one shared parameter is swept.

The results highlight that SF for the GVFs is critical for effective learning, though GPI and Sarsa perform similarly, as shown in Figure \ref{fig:control_tmaze_rmse}. The utility of SF is even greater here, with TB GVF learners inducing much worse performance than SF GVF learners. GPI and Sarsa are similar, which is likely due to the fact that Sarsa uses traces with tabular features, which allow states along the trajectory to the drifter goal to update quickly. In following sections, we find a bigger distinction between the two.

We visualize the goal visitation of GPI in Figure \ref{fig:visitation_tmaze}. Once the GVF learners have a good estimate for the \textit{constant} cumulant signals and the \textit{distractor} cumulant signal, the agent behavior should switch to visiting only the \textit{drifter} cumulant as that is the only goal where visiting would improve the GVF prediction. When using SF GVF learners, this behavior emerges, but under TB GVF learners the agent incorrectly focuses on the distractor. This is even more pronounced for Sarsa (see Appendix~\ref{app:sarsa-tmaze}). 

\section{Experiments under Function Approximation}

We evaluate our system in a similar fashion to the last section, but now under function approximation. We use a benchmark problem at the end of this section, but start with experiments in the TMaze modified to be a continuous environment, with full details described in Appendix~\ref{app:tmazedetails}. The environment observation $o_t \in \mathbb{R}^2$ corresponds to the xy coordinates of the agent. We use tile coded features of 2 tilings of 8 tiles for the state representation, both for TB and to learn the SF. 

The reward features for the GVF learners can be much simpler than the state-action features because they only need to estimate the cumulants, which are zero in every state except the goals. The reward features are a one-hot encoding indicating if $s'$ in the tuple of $(s,a,s')$ is in the pseudo-termination goals of the GVFs. For the Continuous TMaze, this gives a 4 dimensional vector. The reward features for GPI is state aggregation applied along the one dimensional line components. 
% MARTHAC: I don't understand this, an n is generic, so omitting
%It is further subdivided into $n$ equal length subparts. 
Appendix~\ref{app:tmazedetails} contains more details on the reward features for the GVF and behavior learners. 

\textbf{Results for a Fixed Behavior and Learned Behaviors}

Under function approximation, SF-NR continues to enable more effective tracking of the cumulants than the other methods. 
For control, GPI is notably better than Sarsa, potentially because under function approximation eligibility traces are not as effective at sweeping back changes in behavior rewards and so the separation is more important. We include visitations plots in Appendix~\ref{app:gpi-tmaze}, which are similar to the tabular setting. 
%Figure \ref{fig:visitation} shows that  GPI quickly identifies the drifting cumulant and repeatedly visits that area of the TMaze for both TB and SF-NR GVF learners. Sarsa, on the other hand, was unable to converge on visiting the drifter with TB GVF learners and only somewhat favoured the drifting cumulant with SF-NR GVF learners.
 
%  \begin{figure}[t]
% 	\begin{subfigure}[b]{0.31\textwidth}
%         \centering
%         \includegraphics[width=\textwidth]{figures/1d_tmaze/RMSE_oned_tmaze_dmu_error.pdf}
%         \caption{Fixed-Behavior Policy}
%         \label{fig:round_robin_rmse}
%     \end{subfigure}
% 	\begin{subfigure}[b]{0.31\textwidth}
%         \centering
%         \includegraphics[width=\textwidth]{figures/1d_tmaze/RMSE_oned_tmaze_error.pdf}
%         \caption{Learned Behavior Policy}
%         \label{fig:control_tmaze_rmse}
%     \end{subfigure}     
% 	\begin{subfigure}[b]{0.37\textwidth}
%         \centering
%         \includegraphics[width=\textwidth]{figures/1d_tmaze/goal_visitation_gpi.png}
%         \caption{Visitation Plot Comparison}
%         \label{fig:visitation}
%     \end{subfigure}      
%     \caption{Performance in \textbf{Continuous TMaze}, with averages over 30 runs. \textbf{(a)} and \textbf{(b)} show average off-policy prediction RMSE, with standard errors, where the error is weighted by (a) the state distribution $\bdist$ for the Fixed-Behavior policy and (b) a uniform state weighting when learning the behavior. 
%     \textbf{(c)} Goal visitation plots for GPI with SF and TB.}
%\end{figure}
 \begin{figure}[t]
 	\begin{subfigure}[b]{0.31\textwidth}
         \centering
         \includegraphics[width=\textwidth]{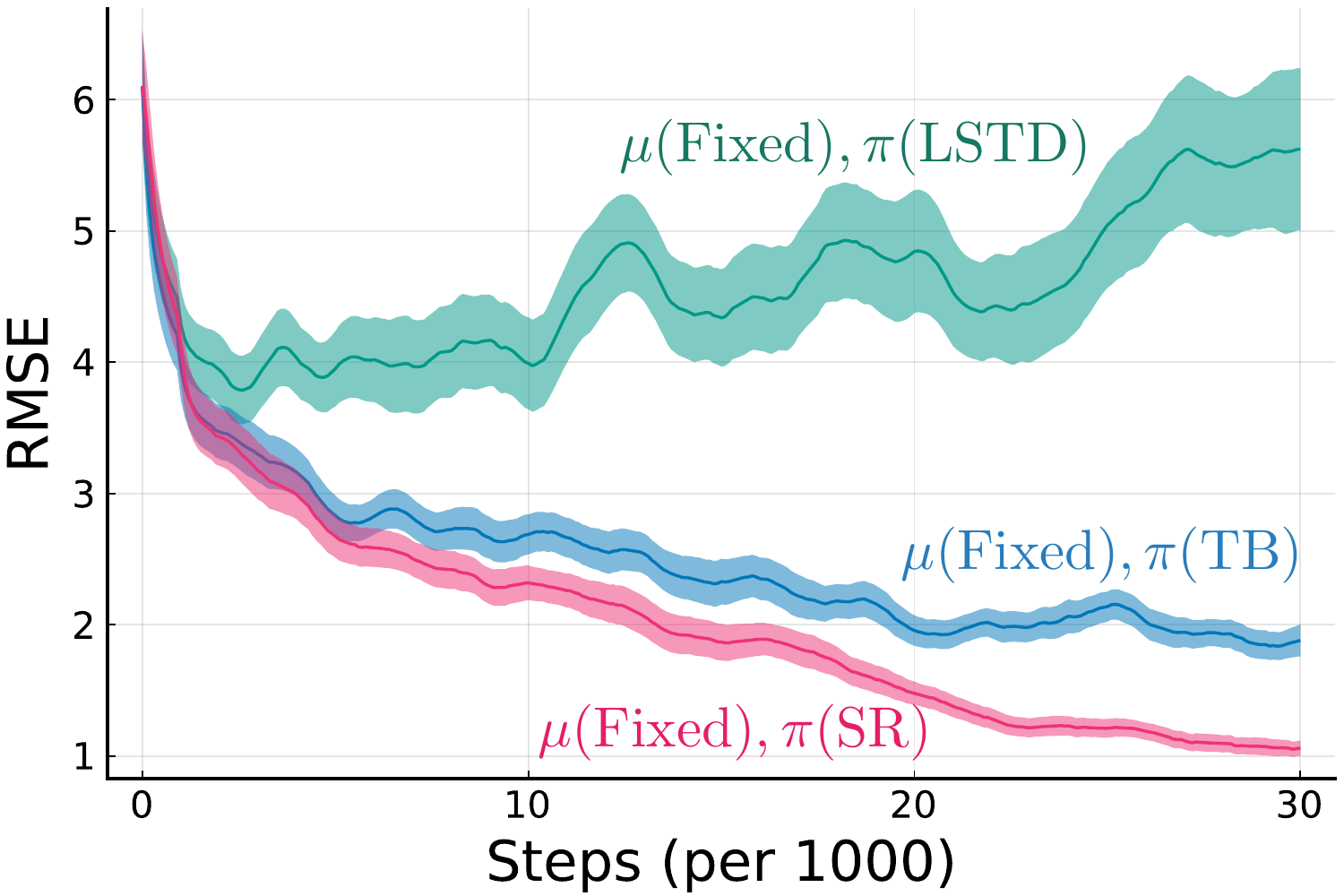}
         \caption{Fixed-Behavior Policy}
         \label{fig:round_robin_rmse}
     \end{subfigure}
 	\begin{subfigure}[b]{0.31\textwidth}
         \centering
         \includegraphics[width=\textwidth]{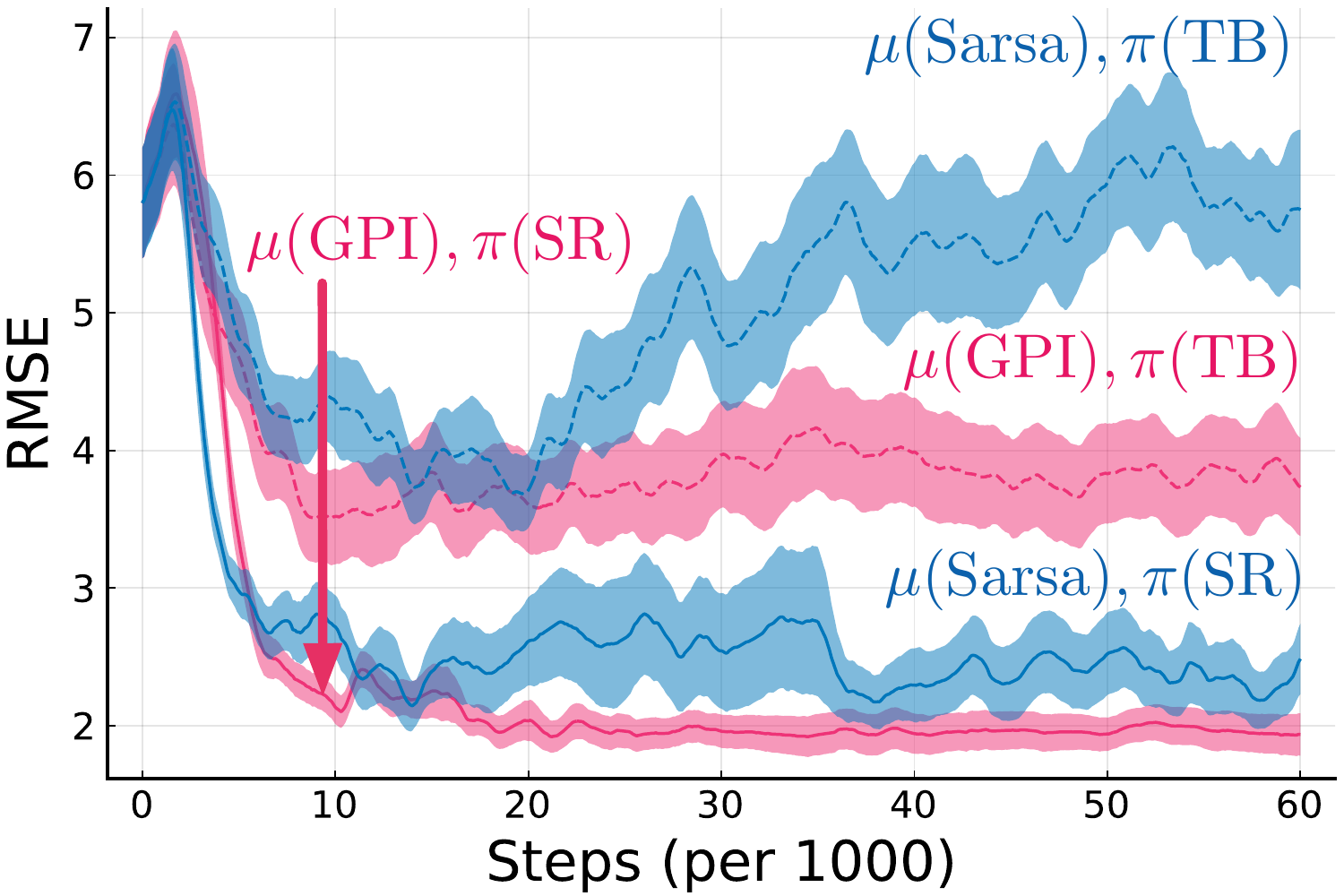}
         \caption{Learned Behavior Policy}
         \label{fig:control_tmaze_rmse}
     \end{subfigure}     
 	\begin{subfigure}[b]{0.31\textwidth}
         \centering
         \includegraphics[width=\textwidth]{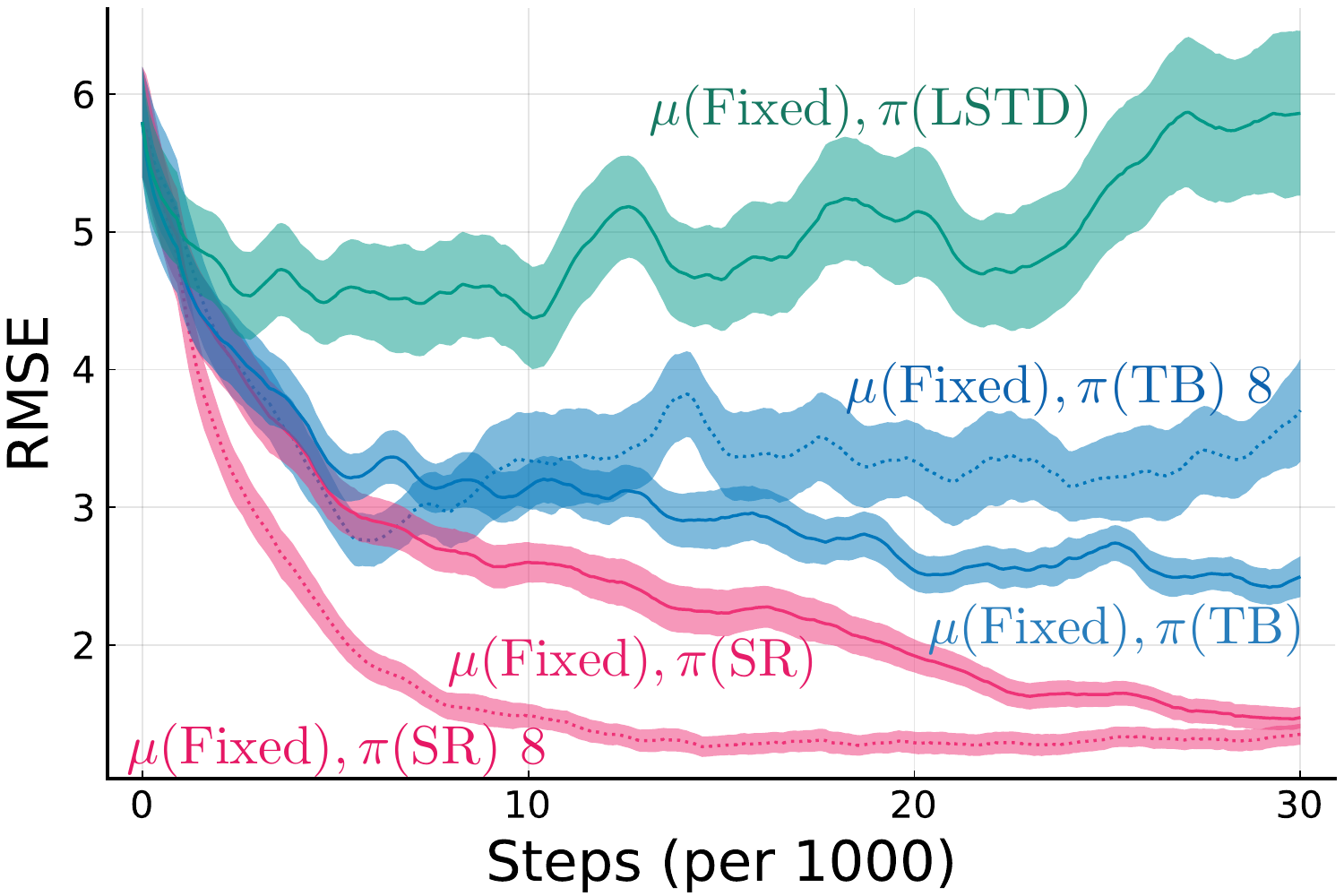}
         \caption{Replay with Fixed-Behavior}
         \label{fig_replay}
     \end{subfigure}      
     \caption{Performance in \textbf{Continuous TMaze}, with averages over 30 runs. \textbf{(a)} and \textbf{(b)} show average off-policy prediction RMSE, with standard errors, where the error is weighted by (a) the state distribution $\bdist$ for the Fixed-Behavior policy and (b) a uniform state weighting when learning the behavior. 
     \textbf{(c)} RMSE in Continuous TMaze with a Fixed Behavior when incorporating replay.}
          \vspace{-0.4cm}
\end{figure}

Note that the efficacy of SF-NR and GPI relied on having reward features that did not overly generalize. The SF learns the expected feature vector when following the target policy. For the GVF learners, if states on the trajectory share features with states near the goal, then the value estimates will likely be higher for those states. The rewards are learned using squared error, which unlike other losses, is likely only to bring cumulant estimates to near zero. These small non-zero cumulant estimates are accumulated by the SF for the entire trajectory, resulting in higher error than TB. We demonstrate this in Appendix~\ref{app:reward-features}. We designed reward features to avoid this problem for our experiments, knowing that effective reward features can and have been learned for SF \citep{barreto2020fast}. 

\textbf{Results using Replay}

%\begin{wrapfigure}[13]{r}{0.4\textwidth}
%\vspace{-0.3cm}
%  \begin{centering}
%      \includegraphics[width=0.4\textwidth]{figures/1d_tmaze/Experience_Replay_RR.png}
%  \end{centering}
%  \vspace{-0.4cm}
%    \caption{RMSE in Continuous TMaze with a Fixed Behavior when incorporating replay. }\label{fig_replay}  
%\end{wrapfigure}
The above results uses completely online learning, with eligibility traces. A natural question is if the more modern approach of using replay could significantly change the results. In fact, early versions of the system included replay but had surprisingly negative results, which we later realized was due to the inherent non-stationarity in the system. Replaying old cumulants and rewards, that have become outdated, actually harms performance of the system. Once we have the separation with the SF, however, we can actually benefit from replay for this stationary component. 
%
%\begin{figure}[h]
%    \includegraphics[width=8cm]{figures/1d_tmaze/Experience_Replay_RR.png}
%    \centering
%    \caption{\textbf{Result with and without replay.} Across all variations tested, the use of replay did not decrease the RMSVE. Performance especially decreased when replay was used to improve both the successor features and reward estimate with SF-NR.}
%    \label{fig:replayexperiment}
%\end{figure}

We demonstrate this result in Figure \ref{fig_replay}. We use $\lambda = 0$ for this result, because we use replay. The settings are otherwise the same as above, and we resweep hyperparameters for this experiment. SF-NR benefits from replay, because it only uses it for its stationary component: the SF. TB, on the other hand, actually performs more poorly with replay. As before, LSTD which similarly uses old cumulants, also performs poorly.

\textbf{Incorporating Interest}
%Due to space constraints, we include results investigating the impact of interest in Appendix~\ref{app:interest-funcapprox}. Those results show the utility of focusing the state-action weighting for the GVF learners. 

\begin{wrapfigure}[10]{r}{0.34\textwidth}
\vspace{-0.75cm}
 	\includegraphics[width=0.34\textwidth]{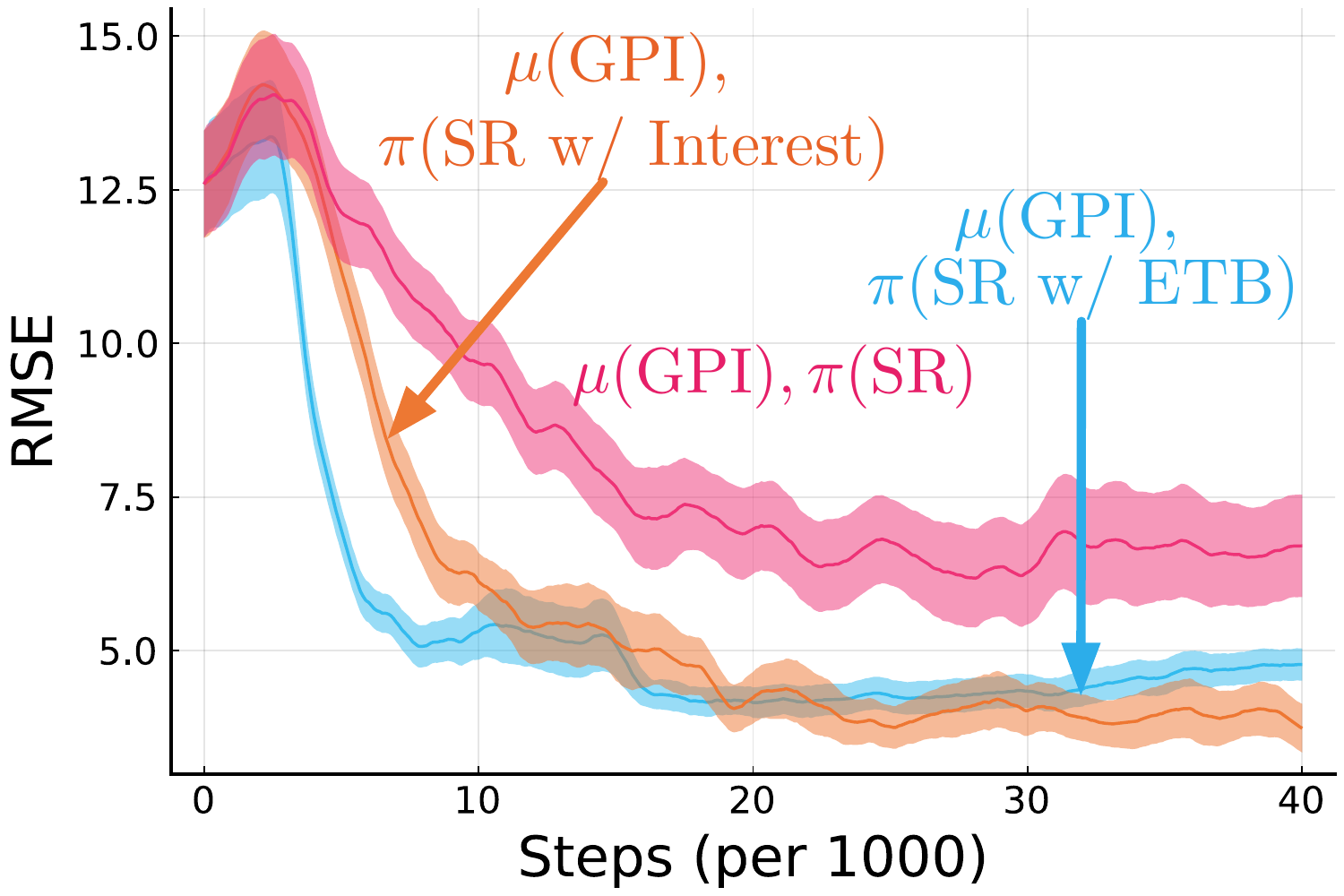}
 	\caption{Using interest: shading is standard error over 30 runs.}
 	\label{fig:state_reweighting_rmse}
\end{wrapfigure}
To study the effects of interest, a more open world environment is needed. The \emph{Open 2D World} is used to analyze this problem as described in Appendix  \ref{app:2dworlddetails}. At the start of each episode, the agent begins in the center of the environment. The interest for each GVF in the states is one if the state is in the same quadrant as the GVF's respective goal, and zero otherwise. This enables the GVFs to focus their learning on a subset of the entire space and thus use the function approximation resources more wisely and give a better weight change profile as an intrinsic reward to the behavior learner. Each GVF prediction $i$ is evaluated under state-action weighting induced by running $\pi_i$, with results in Figure \ref{fig:state_reweighting_rmse}. 
%The corresponding state distribution for the ith GVF weighting for RMSE calculation is the $d_{\pi_i}$ induced by following policy $\pi_i$ from the start state distribution. The algorithms are selected based on lowest overall RMSE on the last 10\% of the runs. The use of interest can help GVF learners in learning about a subset of an MDP.

Both TB with interest and ETB reweight states to focus more on state visitation under the policy. Both significantly improve performance over not using interest, both allowing faster learning and reaching a lower error. The reweighting under ETB more closely matches state visitation under the policy, and accounts for the impacts of bootstrapping. We find that ETB does provide some initial learning benefits. The original ETD algorithm is known to suffer from variance issues; we may find with variance reduction that the utility of ETB is even more pronounced.  
%However, the original algorithm ETD is known to suffer from variance issues. Here, we find ETB does not provide improvements over TB with interest, likely for this reason. With improvements to using algorithms like ETD, we expect we can improve on the performance of just incorporating interest.

% \begin{figure}[h]
% 	\centering
% 	\includegraphics[width=0.6\textwidth]{figures/state_reweighting/TwoD_RMSE.pdf}
% 	\caption{GPI with PCG where the GVFs are given interest in states near their goal are evaluated on the state distribution induced by $d_{\pi_i}$. Shaded region is the standard error across 30 runs.}
% 	\label{fig:state_reweighting_rmse}
% \end{figure}

\textbf{Validation of the Multi-Prediction System in a Standard Benchmark Problem}

Finally, we investigate multi-prediction learning in an environment not obviously designed for this setting: Mountain Car. The goal here is to show that multi-prediction learning is natural in many problem settings, and to show results in a standard benchmark not designed for our setting that has more complex transition dynamics.  
%The previous experiments focused on navigation environments designed for multi-prediction learning; does our approach extend to other tasks with more complex dynamics? To investigate this we used a classic 2D continuous state control task, Mountain Car. 
In the usual formulation the agent must learn to rock back and forth building up momentum to reach the top of the hill on the right---a classic cost to goal problem. This is a hard exploration task where a random agent requires thousands of steps to reach the top of the hill from the bottom of the valley. Here we use Mountain Car to see if our approach can learn about more than just getting out of the valley quickly. We specified a GVF whose termination and policy focuses on reaching top of the left hill, and a second GVF about reaching the top of the other side. The full details of the GVFs and setup of this task can be found in the Appendix~\ref{app:mcdetails}.

Figure \ref{fig:mc_rmse} shows how GPI and Sarsa compare against a baseline random policy. GPI provides much better data for GVF learning than the random policy and Sarsa, significantly reducing the RMSE of the learned GVFs. The goal visitation plots show GPI explores the domain and visits both GVFs goal far more often than random, and more effectively than Sarsa.   

%\begin{figure}[h]
%    \includegraphics[width=6cm]{figures/mountain_car/RMSE_mc_uniform_error.pdf}
 %   \centering
  %  \caption{Shows the average off-policy RMSE error of the two demons weighted by the state distribution weighted uniformly across all valid states and actions}
   % \label{fig:mc_rmse}
%\end{figure}

 \begin{figure}[t]
	\begin{subfigure}[b]{0.33\textwidth}
		\centering
		\includegraphics[width=\textwidth]{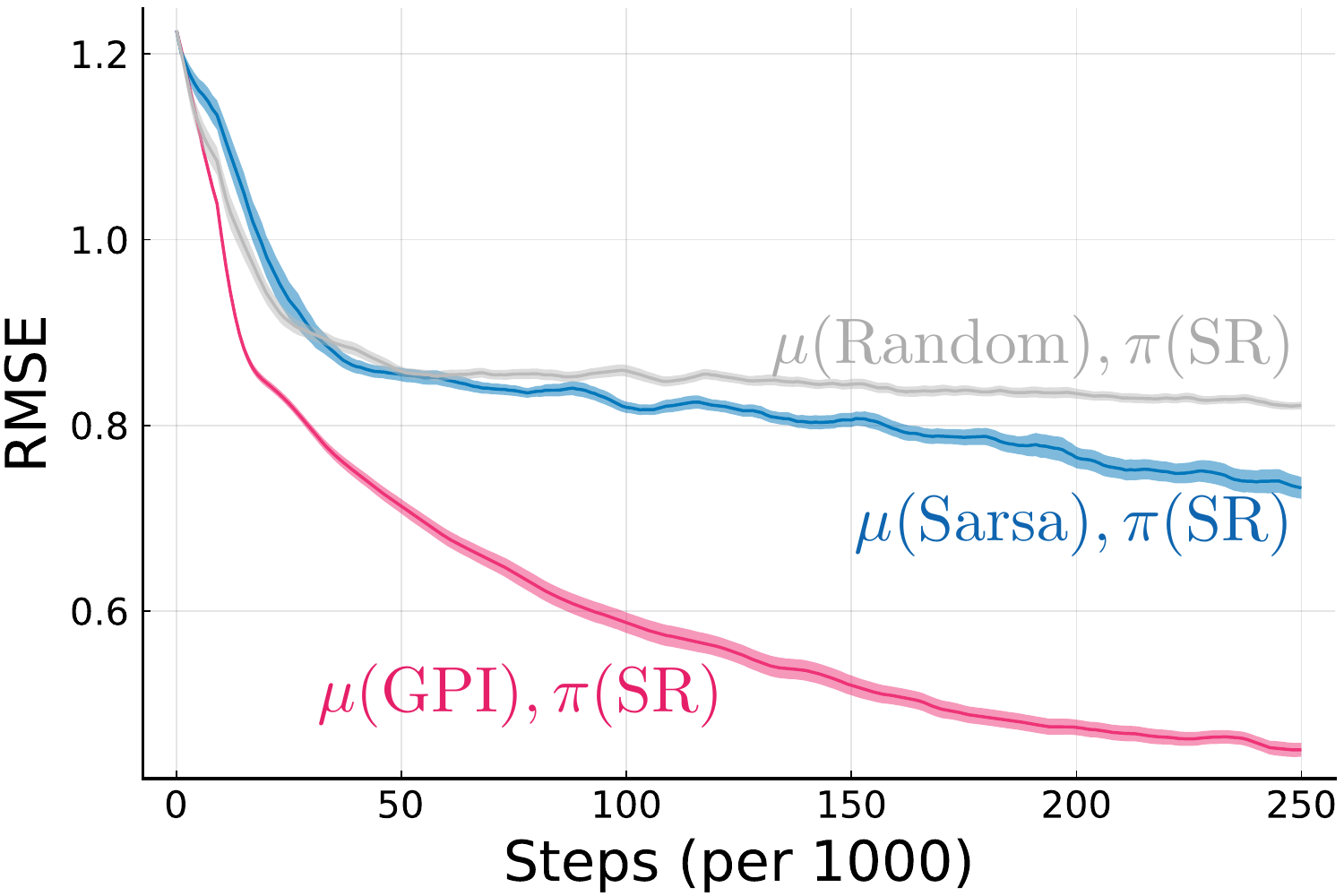}
		\caption{RMSE for each GVF}
		\label{fig:mc_rmse}
	\end{subfigure}
	\begin{subfigure}[b]{0.33\textwidth}
		\centering
		\includegraphics[width=\textwidth]{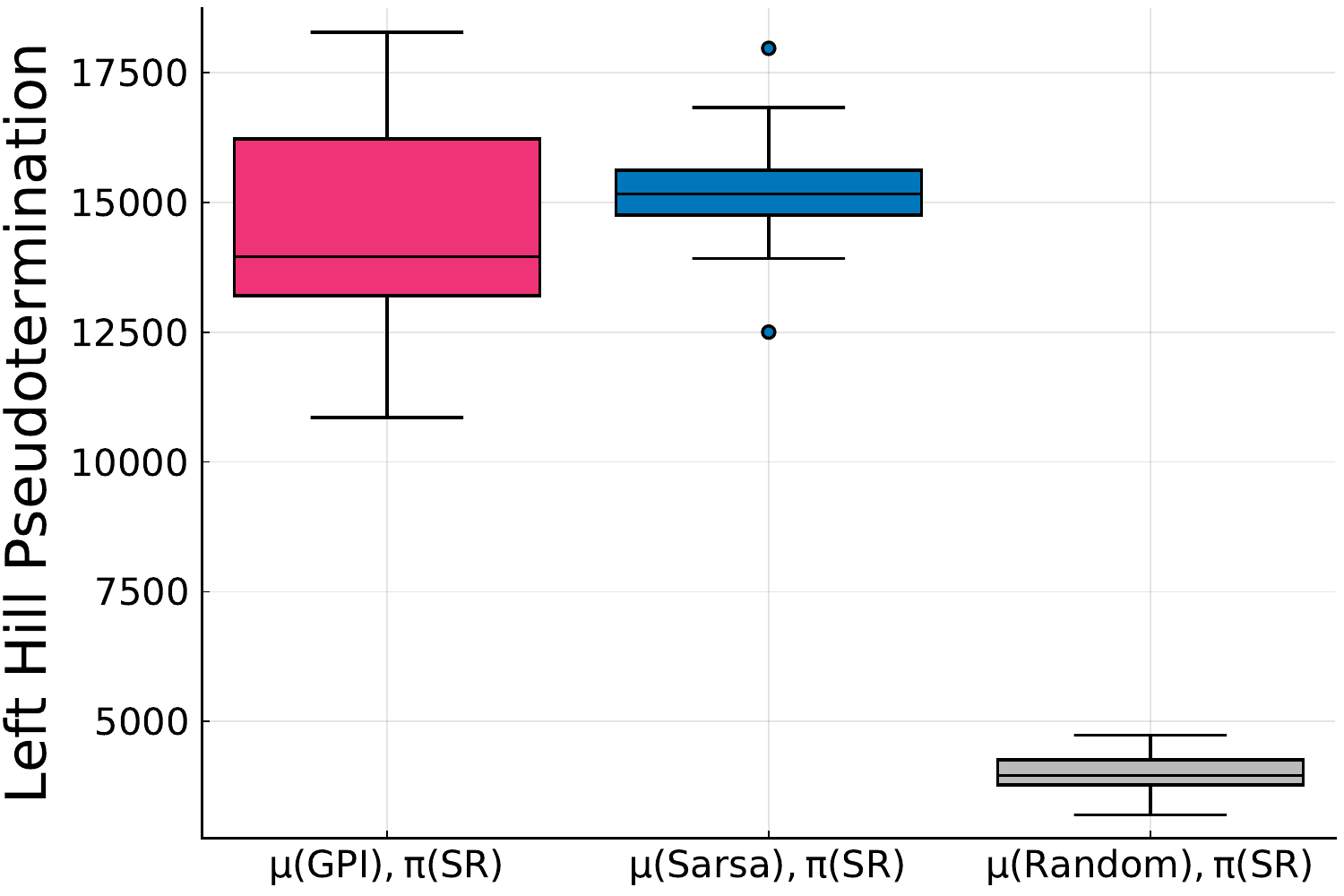}
		\caption{GVF\#2 goal visits: left hill top}
		\label{fig:right_wall_terms}
	\end{subfigure}     
	\begin{subfigure}[b]{0.33\textwidth}
		\centering
		\includegraphics[width=\textwidth]{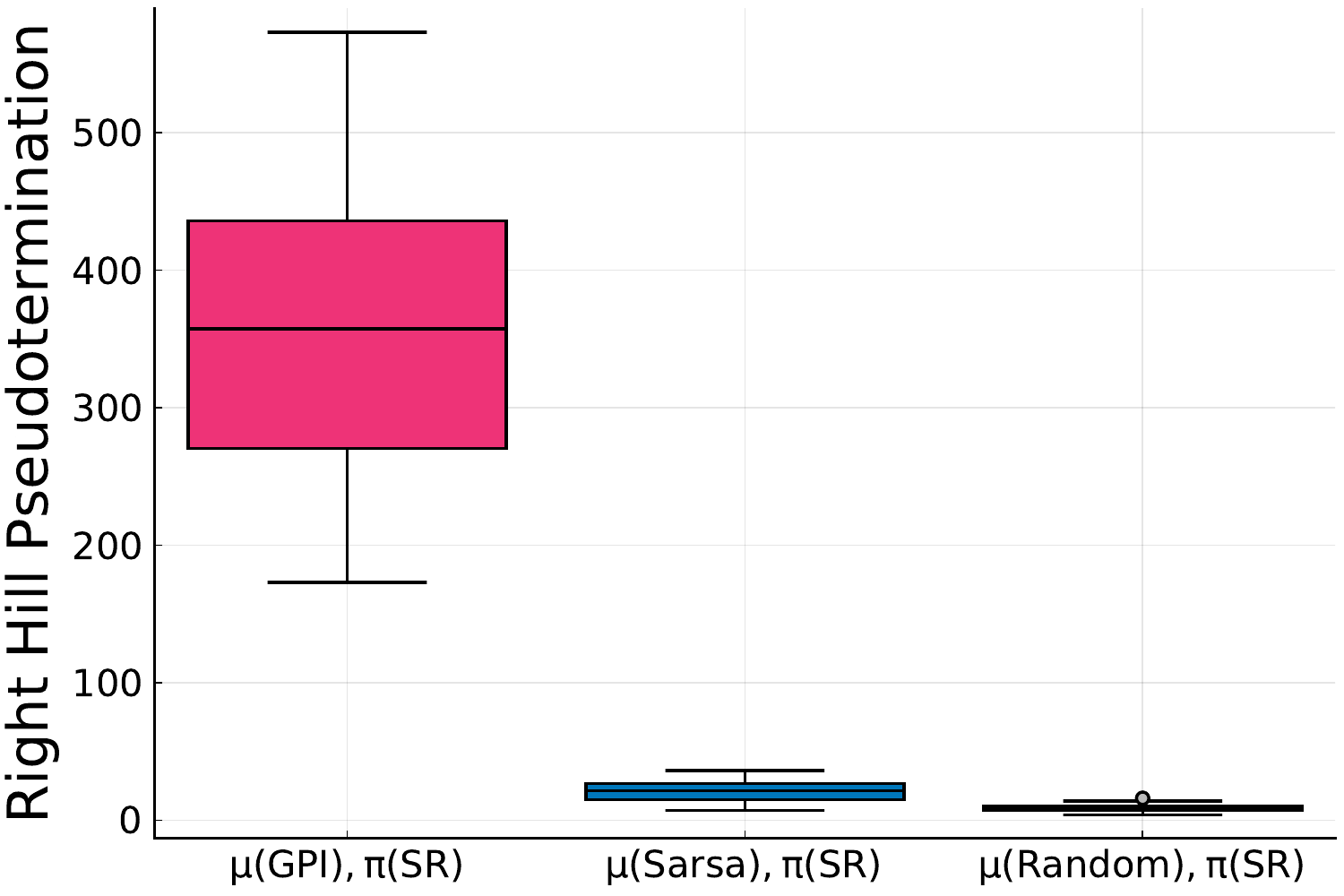}
		\caption{GVF\#1 goal visits: right hill top}
		\label{fig:hill_terms}
	\end{subfigure}      
	\caption{Performance in \textbf{Mountain Car} averaged over 30 runs, with standard errors. \textbf{(a)} Learning curves for RMSE, with a uniform weighting over states and actions. 
	%average off-policy prediction RMSE (uniform weighted over states and actions), with standard errors. 
	 \textbf{(b)}, \textbf{(c)} show the number of times that the agent reached the termination for each GVF.}
	 \vspace{-0.6cm}
\end{figure}

%\textbf{Learning A Complex Behaviour From Intrinsic Rewards}
%
%
%GPI with SF-NR enables the agent to craft more complex policies than ESarsa. This is indicated in Figure \ref{fig:hill_terms} where GPI is able to succesfully navigate to more difficult to reach regions of the state space such as the termination zone on the right hill. Reaching this area requires a policy of rocking back and forth to build momentum which is unlikely to occur by chance (as demonstrated by the agent following a random behaviour policy). 
%
%\Matthew{This needs some kind of concluding sentence I think}

\vspace{-.2cm}

\section{Conclusion}
\label{sec:conclusion}

In this work, we take the first few steps towards building an effective multi-prediction learning system. We highlight the inherent non-stationarity in the problem and design algorithms based on successor features (SF) to better adapt to this non-stationarity. 
%We provide several diagnostic experiments highlighting 
We show that 
(1) temporally consistent behavior emerges from optimizing the amount of learning across diverse GVF questions; 
(2) successor features are useful for tracking nonstationary rewards and cumulants, both in theory and empirically;
%(2) in addition to zero-shot transfer, SFs are an effective way to address learning non-stationary signals in both prediction and control; 
(3) replay is well suited for learning the stationary components successor features while meta-learning works well for the non-stationary components; 
and (4) interest functions can improve the performance of the entire system, by focusing learning to a subset of states for each prediction.
% THis is more of an open question rather than a takeway
%(5) specialized learning algorithms for the behaviour may be needed to address the non-stationary intrinsic reward stream. 

Our work also highlights several critical open questions. (1) The utility of SFs is tied to the quality of the reward features; better understanding of how to learn these reward features is essential. (2) Continual Auxiliary Task Learning is an RL problem, and requires effective exploration approaches to find and maximize intrinsic rewards---the intrinsic rewards do not provide a solution to exploration. Never-ending exploration is needed. (3) The interaction between discovering predictive questions and learning them effectively remains largely unexplored. In this work, we focused on learning, for a given set of GVFs. Other work has focused on discovering useful GVFs \citep{veeriah2019discovery,veeriah2021discovery, nair2020contextual,zahavy2021discovering}. The interaction between the two is likely to introduce additional complexity in learning behavior, including producing automatic curricula observed in previous work \citep{oudeyer2007intrinsic, chentanez2005intrinsically}. 

%Throughout this work, we encountered issues that remain open problems for further investigation. The intrinsic reward indicates how the agent should act in the absence of external rewards and not how the agent should explore the environment. Exploration techniques to find the states that maximize the amount of learning remains a complementary research question. We also encountered sensitivity of the reward features of SF on agent performance. Further study into learning the reward representation may be necessary for more challenging problems. Lastly, how the GVF questions can be discovered remains an open problem that synergizes with this work. By maximizing the amount of learning, the agent will likely generate its own curriculum with learning the easier auxiliary tasks first. By having the agent create its own auxiliary tasks, it helps ensure the agent has a rich intrinsic reward to guide behaviour throughout the agent's lifetime. Correspondingly, by learning the auxiliary tasks more accurately, the downstream use cases of learning auxiliary tasks may be improved which potentially can include the discovery of auxiliary tasks themselves \citep{veeriah2019discovery,veeriah2021discovery, nair2020contextual}. 

This work demonstrates the utility of several new ideas in RL that are conceptually compelling, but not widely used in RL systems, namely SF and GVFs, GPI with SF for control, meta-descent step-size adaption, and interest functions. 
The trials and tribulations that lead to this work involved many failures using classic algorithms in RL, like replay; and, in the end, providing evidence for utility in these newer ideas. Our journey highlights the importance of building and analyzing complete RL systems, where the interacting parts---with different timescales of learning and complex interdependencies---necessitate incorporating these conceptually important ideas. Solving these integration problems represents the next big step for RL research.

\begin{ack}
This work was supported by NSERC Discovery, IVADO, CIFAR through CCAI Chair funding and by the Alberta Machine Intelligence Institute (Amii). 
\end{ack}

\bibliographystyle{plainnat} % We choose the "plain" reference style
\bibliography{refs} % Entries are in the "refs.bib" file

 \newpage
\appendix
\newcommand{\Amatpi}{\mathbf{A_\pi}}
\newcommand{\Amatinv}{\mathbf{A_{inv}}}
\newcommand{\Pimat}{\mathbf{\Pi_\pi}}
\newcommand{\Pgmat}{\mathbf{P_\gamma}}
\newcommand{\Projmat}{\mathbf{\Pi}}
\newcommand{\Psimat}{\mathbf{\Psi}}

\newcommand{\bvecpi}{\mathbf{b_\pi}}
\newcommand{\etavecr}{\mathbf{\boldsymbol{\eta}_r}}
\newcommand{\wvecpi}{\wvec^\pi}

\newcommand{\thetavec}{\mathbf{\boldsymbol{\theta}}}
\newcommand{\alphavec}{\mathbf{\boldsymbol{\alpha}}}

\section{Sample Efficiency of SF-NR} \label{app:sample-eff}
\subsection{Proof of Lemma~\ref{lemma:msve-bound}}\label{app:msve-bound}
In this section we provide a proof of Lemma~\ref{lemma:msve-bound}. The result is included for completeness, and
similar results of a similar form can be found throughout the literature (for example, similar steps are used in the
proof of Lemma 3 of \citet{scherrer2016improved}). The lemma is repeated for convenience below.

\MSVEBound*

\textit{Proof:}
Given access to
an estimate $\hat \rvec $ of $\rvec ^{\pi}$ can then bound the
MSVE as
\begin{align}
  \half\norm{\vvec^{\pi}-\hat \vvec}_{\Dmat}^{2}
  &\overset{(a)}{=} \half\norm{(\eye-\gamma \Pmat_{\pi})^{\inv}\brac{\rvec ^{\pi} - \hat \rvec }}^{2}_{\Dmat}\nonumber\\
  &\overset{(b)}{\le}\half\norm{(\eye-\gamma \Pmat_{\pi})^{\inv}}^{2}_{\Dmat}\norm{\rvec ^{\pi}-\hat \rvec }^{2}_{\Dmat}\nonumber\\
                          &\overset{(c)}{\le}\half\brac{\sum_{t=0}^{\infty}\gamma^{t}\norm{\Pmat_{\pi}^{t}}_{\Dmat}}^{2}\norm{\rvec ^{\pi}-\hat \rvec }^{2}_{\Dmat}\nonumber\\
  &\overset{(d)}{\le}\frac{\norm{\rvec ^{\pi}-\hat \rvec }_{\Dmat}^{2}}{2(1-\gamma)^{2}}\nonumber
  % \\
  % &=\half\frac{\E_{s\sim d}\sbrac{\brac{r^{\pi}(s)-\hat r(s)}^{2}}}{(1-\gamma)^{2}}\label{eq:vrate},
\end{align}
where $(a)$ decomposed $\vvec^{\pi}=(\eye-\gamma \Pmat_{\pi} )^{\inv}\rvec^{\pi}$,
$(b)$ uses sub-multiplicativity of the matrix norm induced by $\norm{\cdot}_{D}$,
$(c)$ uses the von-neumann expansion $(\eye-\gamma \Pmat_{\pi})^{\inv}=\sum_{t=0}^{\infty}\gamma^{t}\Pmat_{\pi}^{t}$ and triangle inequality, and
$(d)$ uses that $\norm{\Pmat_{\pi}^{t}}_{D} =\lambda_{\max{}}\brac{(\Pmat_{\pi}^{t})^{\top}\Dmat\Pmat_{\pi}^{t}}\le 1$ since both $\Dmat$ and $\Pmat_{\pi}^{t}$ have eigenvalues
of at-most 1, followed by $\sum_{t=0}^{\infty}\gamma^{t}=\frac{1}{1-\gamma}$.

\hfill$\blacksquare$

\subsection{Proof of Proposition~\ref{prop:regret-bound}}\label{app:regret-bound}

\RegretBound*

\textit{Proof:}
\begin{align*}
  \cL(\wbar_{T}) &= \EE_{d_{\pi}}\sbrac{\half\brac{r_{t}-\inner{x(S_{t})}{\wbar_{T}}}^{2}}\\
  &=\EE_{\bdist}\sbrac{\frac{\rho_{t}}{2}\brac{r_{t}-\inner{x(S_{t})}{\wbar_{T}}}^{2}}\\
  &\le\EE_{\bdist}\sbrac{\frac{1}{T}\sum_{t=1}^{T}\ell_{t}(w_{t})}\\
  &\le\frac{\norm{w^{*}}^{2}+d \rho_{\max{}}R_{\max{}}^{2}\Log{1+\rho_{\max}L^{2}T}}{2T},
\end{align*}
where in the last line we applied the regret guarantee of the RLS estimator with regularization parameter $\lambda=1$ (See \citet[Theorem 7.26]{orabona2019modern})
and used that $\max_{s}\norm{\phi(s)}_{2}\le\sqrt{d}\max_{s}\norm{\phi(s)}_{\infty}=\sqrt{d}L$.
% \AJ{ the log term is actually $\Log{1+\frac{\rho_{\max}(\widetilde{L})^{2}T}{d}}$ where $\widetilde{L}=\max\norm{\phi(s)}_{2}$ in \citet{orabona2019modern}. I
%   upperbounded this with $\widetilde{L}\le\sqrt{d}L=\sqrt{d}\max_{s}\norm{\phi(s)}_{\infty}$ to better-align with \citet{liu2018proximal}.
% }
Following Lemma \ref{lemma:msve-bound}, by taking $\hat v(s) = \inner{\psi(s)}{\wbar_{T}}$, we have that
\begin{align}
  \norm{\vvec^{\pi}-\hat\vvec}^{2}_{\Dmat}
  %&\qquad\le \frac{\norm{w^{*}}^{2}+dR_{\max{}}^{2}\Log{1+\frac{\rho_{\max{}}L^{2}T}{d}}}{2(1-\gamma)^{2}T}\\
  \le O\brac{\frac{d \rho_{\max}R_{\max{}}^{2}\Log{1+\rho_{\max{}}L^{2}T}}{(1-\gamma)^{2}T}}\label{eq:sf-bound-app}.
\end{align}
\hfill$\blacksquare$

%\clearpage
\section{Relationship between SF-NR and TD solutions}
\label{app:relationship-srnf-td}

Let $\wvecpi$ be the fixed-point solution for the projected Bellman operator with respect to the $\lambda$-return that is be estimated by LSTD($\lambda$) for policy $\pi$\citep{white2017unifying}. It is well known that TD($\lambda$) converges to this solution under the right conditions.
The components of the solution $\wvecpi = \Amatpi^\inv \bvecpi$ are as follows
\begin{align*}
\Amatpi &= \Xmat^\top \Dmat (\eye - \lambda \Pgmat \Pimat)^\inv (\eye - \Pgmat \Pimat)\Xmat \\
\bvecpi &= \Xmat^\top \Dmat (\eye - \lambda \Pgmat \Pimat)^\inv \rvec
\end{align*}
where $\Xmat\in \RR^{|\States||\Actions|\times\numfeats}$ is the feature matrix with $\xvec(s,a)^\top$ along its rows, $\rvec \in \RR^{|\States||\Actions|}$ is the expected immediate reward  \big($\rvec((s,a))=\sum_{s'\in\States} P(s,a,s') R(s,a,s')$\big),
$\Pmat_\gamma \in \RR^{|\States||\Actions|\times|\States|}$ is a sub-stochastic matrix that represents the transition process 
\big($\Pmat_\gamma((s,a),s')= P(s,a,s')\gamma(s,a,s')$\big), $\Pimat \in \RR^{|\States| \times |\States||\Actions|}$ is a stochastic matrix that represents $\pi$ \big($\Pimat(s,(s,a)\big) = \pi(s,a)$), and $\Dmat \in \RR^{|\States| \times |\States||\Actions|}$ is a diagonal matrix with the stationary distribution induced by $\pi$ on its diagonal that controls the approximation error.

Let us consider $\lambda=1$ case for simplicity. Under this case 
\begin{align*}
\Amatpi &= \Xmat^\top \Dmat \Xmat \\
\bvecpi &= \Xmat^\top \Dmat (\eye - \Pgmat \Pimat)^\inv \rvec
\end{align*}
The predicted values correspond to $\hat{\Qmat} = \Xmat \wvecpi$. This is the projection of the true values $\Qmat^* = (\eye - \Pgmat \Pimat)^\inv \rvec$ onto the space spanned by $\Xmat$ where the projection operator is $\Projmat = \Xmat (\Xmat^\top \Dmat \Xmat)^\inv \Xmat^\top \Dmat$ -- the TD(1) solution. Now, depending on if the form of the reward $\rvec$ we have the three following cases.

\textit{Case 1: $\rvec = \Xmat \wvec$} The LSTD estimate can be written as 
\begin{align*}
\theta_{\text{LSTD}} = (\Xmat^\top \Dmat \Xmat)^\top \Xmat^\top \Dmat (\eye - \Pgmat \Pimat)^\inv \Xmat \wvec
\end{align*}
where the component $\Psimat=(\eye - \Pgmat \Pimat)^\inv \Xmat$ corresponds to the successor features.
Therefore, if the space $\Xmat$ is used for learning both the successor features and the reward, the solution corresponding to SF-NR would be equivalent to the solution obtained by TD.

\textit{Case 2: $\rvec = \Phimat \wvec$} The LSTD estimate can be written as 
\begin{align*}
\theta_{\text{LSTD}} = (\Xmat^\top \Dmat \Xmat)^\top \Xmat^\top \Dmat (\eye - \Pgmat \Pimat)^\inv \Phimat \wvec
\end{align*}
where the component $\Psimat=(\eye - \Pgmat \Pimat)^\inv \Phimat$ corresponds to the successor features.
Therefore, if the space $\Xmat$ is used for learning the successor features which correspond to weighted sums of $\Phimat$, and $\Phimat$ is used for learning the reward, the solution corresponding to SF-NR would be equivalent to the solution obtained by TD.

\textit{Case 3: $\rvec = \Xmat \wvec + \etavecr$ or $\rvec = \Phimat \wvec + \etavecr$} where $\etavecr$ is the model misspecification error for predicting the immediate reward, the LSTD estimate would correspond to
\begin{align*}
\theta_{\text{LSTD}} = (\Xmat^\top \Dmat \Xmat)^\top \Xmat^\top \Dmat (\eye - \Pgmat \Pimat)^\inv \rvec
\end{align*}
whereas the SF-NR estimate would capture the same component as in case (1), or in case (2).
Therefore, if there is a misspecification error for learning $\rvec$, the two solutions would differ. 

Hence, decomposition of SF-NR does not reduce representability of the TD(1) solution if the reward is linearizable in some features. More generally, we could introduce $\lambda < 1$ to provide.a bias-variance trade-off for learning the SR as well.

%\raksha{I'm not sure what this means for value-error.. Should we say something about it? I think it should increase for SF-NR compared to TD in the third case because of the orthogonal projection? A case in favour of SF-NR is if SF-NR uses $\Xmat$ and TD uses $\Phimat$ -- assuming $\Xmat$ is a better space for values than $\Phimat$ -- but that seems weird to not use the better space for TD. But alternatively, I guess even though they may differ in expressibility, SF-NR is still more amenable to tracking, which is great?!}

%\clearpage
\section{Prior Corrections and the Projected Bellman Error}
\label{app:prior-correction-pbe}

Let us first consider the SR objective under a fixed behavior, $\bpolicy$, with stationary distribution $\bdist$ over states and actions. When using TD for action-values, with covariance $\Cmat = \mathbb{E}[\xvec(S,A) \xvec(S,A)^\top] = \sum_{s,a} \bdist(s,a)\xvec(S,A) \xvec(S,A)^\top$, the underlying objective is the mean-squared projected Bellman error (MSPBE):
\begin{align*}
\text{MSPBE}(\tweights) 
&=  \| \sum_{s,a} \bdist(s,a) \mathbb{E}_\pi[\delta(\tweights) \xvec(s,a) | S = s, A = a] \|^2_{\Cmat^{-1/2}}\\
&=  \mathbb{E}_\pi[\delta(\tweights) \xvec(S,A)]^\top \Cmat^{-1} \mathbb{E}_\pi[\delta(\tweights) \xvec(S,A)]
\end{align*}   
The TD fixed point corresponds to $\tweights$ such that $\mathbb{E}_\pi[\delta(\tweights) \xvec(S,A)] = 0$, which is defined based on state-action weighting $\bdist$. Different weightings result in different solutions.

The weighting is implicit in the TD update, when updating from state and actions visited under the behavior policy. The predictions are updated more frequently in the more frequently visited state-action pairs, giving them higher weighting in the objective. However, we can change the weighting using important sampling. For example, if we pre-multiply the TD update with $d(s,a)/\bdist(s,a)$ for some weighting $d$, then this changes the state-action weighting in the objective to $d(s,a)$ instead of $\bdist(s,a)$. 

The issue, though, is not that the objective is weighted by $\bdist$, but rather that $\bdist$ is changing as $\bpolicy$ is changing. Correspondingly, the optimal SR solution could be changing since the objective is changing. 
%The level of non-stationarity may not be high, and in some cases is not present. In particular, t
The impact of this changing state distribution depends on the function approximation capacity.
The weighting indicates how to trade-off function approximation error across states; when approximation error is low or zero, the weighting has no impact on the TD fixed point. For example, in a tabular setting, the agent can achieve $\mathbb{E}_\pi[\delta(\tweights) \xvec(s,a) | S = s, A = a] = 0$ for every $(s,a)$. Regardless of the weighting---as long as it is non-zero---the TD fixed point is the same. 
  
Generally, however, there will be some approximation error and so some level of non-stationarity. This pre-multiplication provides us with a mechanism to keep the objective stationary. If we could track the changing $\bdist_t$ with time, and identify a desired weighting $d$, then we could pre-multiply each update with $d(s_t,a_t)/\bdist_t(s_t,a_t)$ to ensure we correct the state-action distribution to be $d$. There have been some promising strategies developed to estimate a stationary $\bdist$ \citep{hallak2017consistent,liu2018breaking,liu2020offpolicy}, though here they would have to be adapted to constantly track $\bdist_t$. 

Another option is to use prior corrections to reweight the entire trajectory up to a state. Prior corrections were introduced to ensure convergence of off-policy TD \citep{precup2000eligibility}. For a fixed behavior, the algorithm pre-multiplies with a product of important sampling ratios, with $\rho(a | s) \defeq \frac{\pi(a | s)}{\bpolicy(a | s)}$
\begin{align*}
\tweights = \tweights + \alpha \left[\Pi_{i=0}^{t}\rho(a_i | s_i) \right]\delta \xvec(s_t, a_t)
\end{align*} 
This shifts the weight from state-actions visited under $\bpolicy$ to state-actions visited under $\pi$, because 
\begin{align*}
&\mathbb{E}_{\bpolicy}\left[\Pi_{i=0}^{t}\rho(A_i | S_i) \delta \xvec(S_t, A_t) | S_t = s, A_t = a \right]\\
  &=\mathbb{E}_{\bpolicy}\left[\Pi_{i=0}^{t}\rho(A_i | S_i) | S_t = s, A_t = a \right] \mathbb{E}[\delta \xvec(S_t, A_t) | S_t = s, A_t = a]
\end{align*}
and when considering expectation across time steps $t$ when $s, a$ are observed
\begin{align*}
\mathbb{E}_{\bpolicy}\left[\Pi_{i=0}^{t}\rho(A_i | S_i) | S_t = s, A_t = a\right] 
&= \frac{d_\pi(s,a)}{\bdist(s,a)}
\end{align*} 

These prior corrections also corrects the state-action distribution even with $\bdist$ changing on each step, because the numerator reflects the probability of reach $s,a$ under policy $\pi$ and the denominator reflects the probability of reach $s,a$ using the sequence of behavior distributions. For $\rho_t(a | s) \defeq \frac{\pi(a | s)}{\bpolicy_t(a | s)}$
\begin{align*}
\Pi_{i=0}^{t}\rho_t(A_i | S_i) 
&= \frac{\pi(A_0 | S_0) \pi(A_1 | S_1) \ldots \pi(A_t | S_t)}{\bpolicy_0(A_0 | S_0) \bpolicy_1(A_1 | S_1) \ldots \bpolicy_t(A_t | S_t)}\\
&= \frac{\pi(A_0 | S_0) P(S_1 | S_0, A_0) \ldots P(S_t | S_{t-1}, A_{t-1}) \pi(A_t | S_t)}{\bpolicy_0(A_0 | S_0) P(S_1 | S_0, A_0) \ldots P(S_t | S_{t-1}, A_{t-1})  \bpolicy_t(A_t | S_t)}
\end{align*} 
%

%\clearpage
\section{Algorithms}\label{app:algs}

The algorithm for Tree-Backup($\lambda$) is from \cite{precup2000eligibility}.

\begin{algorithm}[H]
  \caption{TB($\lambda$) Update}
  \label{alg:TB}
\begin{algorithmic}
  \STATE {$\zvec_t = \gamma_t \pi(A_t|S_t) \lambda \zvec_{t-1} + \xvec(S_t, A_t)$} 
  \STATE{$\delta_t = c_t + \gamma_{t+1} \sum_{a'} \pi(a'|S_{t+1})\estq(S_{t+1}, a')- \estq(S_t, A_t)$}
  \STATE{$\wvec_{t+1} = \wvec_t + \eta_t \delta_t \zvec_t$}
\end{algorithmic}
\end{algorithm}

Algorithm \ref{alg:interestTB} is the online TB with interest update.  The derivation for the online update rule from the forward view is in the next section.
\begin{algorithm}[H]
  \caption{TB($\lambda$) with Interest Update}
  \label{alg:interestTB}
\begin{algorithmic}
  \STATE {$\zvec_t = \gamma_t \pi(A_t|S_t) \lambda \zvec_{t-1} + I_t \xvec(S_t, A_t)$} 
  \STATE{$\delta_t = c_t + \gamma_{t+1} \sum_{a'} \pi(a'|S_{t+1})\estq(S_{t+1}, a')- \estq(S_t, A_t)$}
  \STATE{$\wvec_{t+1} = \wvec_t + \eta_t \delta_t \zvec_t$}
\end{algorithmic}
\end{algorithm}

ETB($\lambda$) is a modified version of ETD($\lambda$) \citep{sutton2016emphatic} using TB($\lambda)$ instead of TD($\lambda)$. This modification relies on the correspondence between TB and TD, where TB is a version of TD with the variable trace parameter, $\lambda_t = b(a_t | s_t) \lambda$ \citep{mahmood2017multi, ghiassian2018online}.
\begin{algorithm}[H]
  \caption{Emphatic TB($\lambda$) Update}
  \label{alg:ETB}
\begin{algorithmic}
  \STATE{$F_t = \rho_{t-1} \gamma_t F_{t-1} + I_t$}
  \STATE{$M_t = \rho_t \Big[\lambda b(A_t| S_t) I_t + \big(1 - \lambda b(A_t|S_t) \big) F_t)\Big]$}
  \STATE {$\zvec_t = \gamma_t \pi(A_t|S_t) \lambda \zvec_{t-1} + M_t \xvec(S_t, A_t)$} 
  \STATE{$\delta_t = c_t + \gamma_{t+1} \sum_{a'} \pi(a'|S_{t+1})\estq(S_{t+1}, a')- \estq(S_t, A_t)$}
  \STATE{$\wvec_{t+1} = \wvec_t + \eta_t \delta_t \zvec_t$}
\end{algorithmic}
\end{algorithm}

%We also include the Emphatic ESARSA($\lambda$) algorithm's update below.
%\begin{algorithm}[H]
%  \caption{Emphatic ESARSA($\lambda$) Update}
%  \label{alg:EmphESARSA}
%\begin{algorithmic}
%  \STATE{$F_t = \rho_{t-1} \gamma_t F_{t-1} + I_t$}
%  \STATE{$M_t = \rho_t \Big[\lambda I_t + \big(1 - \lambda \big) F_t)\Big]$}
%  \STATE {$\zvec_t = \gamma_t \rho_t \lambda \zvec_{t-1} + M_t \xvec(S_t, A_t)$} 
%  \STATE{$\delta_t = c_t + \gamma_{t+1} \sum_{a'} \pi(a'|S_{t+1})\estq(S_{t+1}, a')- \estq(S_t, A_t)$}
%  \STATE{$\wvec_{t+1} = \wvec_t + \eta_t \delta_t  \zvec_t$}
%\end{algorithmic}
%\end{algorithm}

\subsection{Online Interest TB Derivation}

The forward view update that uses interest at each time-step is of the form
\begin{align*}
	\wvec_{t+1} = \wvec_t + \alpha I_t (G_t - \estq(S_t, A_t, \wvec_t)) \nabla \estq(S_t, A_t, \wvec_t).
\end{align*}

According to \citet{suttonbartobook} (page 313), ignoring the changes in the approximate value function, the TB return can be written as, 
\begin{equation}
  G_t \approx \estq(S_t, A_t, \wvec_t) + \sum_{k=t}^\infty \delta_k \prod_{i=t+1}^k \gamma_i \lambda_i \pi(A_i|S_i).
  \label{eq:tb_return}
\end{equation}

We substitute Equation~\ref{eq:tb_return} for $G_t$ in the forward view update, we get,
\begin{align*}
  \wvec_{t+1} \approx \wvec_t + \alpha I_t \sum_{k=t}^\infty \delta_k \prod_{i=t+1}^k \gamma_i \lambda_i \pi(A_i|S_i) \nabla \estq(S_t, A_t, \wvec_t).
\end{align*}

The sum of forward view update over time is
\begin{align*}
\sum_{t=1}^\infty (\wvec_{t+1} - \wvec_t) &\approx  \sum_{t=1}^\infty \sum_{k=1}^\infty \alpha I_t \delta_k \nabla \estq(S_t, A_t, \wvec_t) \prod_{i=t+1}^k \gamma_i \lambda_i \pi(A_i|S_i) \\
&= \sum_{k=1}^\infty \sum_{t=1}^k \alpha I_t \nabla \estq(S_t, A_t, \wvec_t)  \delta_k \prod_{i=t+1}^k \gamma_i \lambda_i \pi(A_i|S_i) \\
&= \sum_{k=1}^\infty \alpha \delta_k \sum_{t=1}^k I_t \nabla \estq(S_t, A_t, \wvec_t)  \prod_{i=t+1}^k \gamma_i \lambda_i \pi(A_i|S_i).
\end{align*}

This can be a backward-view TD update if the entire expression from the second sum can be estimated incrementally as an eligibility trace. Therefore
\begin{align*}
\zvec_k &= \sum_{t=1}^k I_t \nabla \estq(S_t, A_t, \wvec_t)  \prod_{i=t+1}^k \gamma_i \lambda_i \pi(A_i|S_i) \\
&= \sum_{t=1}^{k-1} I_t \nabla \estq(S_t, A_t, \wvec_t)  \prod_{i=t+1}^k \gamma_i \lambda_i \pi(A_i|S_i) + I_k \nabla \estq(S_k, A_k, \wvec_k) \\
&= \gamma_k \lambda_k \pi(A_k|S_k) \sum_{t=1}^{k-1} I_t \nabla \estq(S_t, A_t, \wvec_t)  \prod_{i=t+1}^{k-1} \gamma_i \lambda_i \pi(A_i|S_i) + I_k \nabla \estq(S_k, A_k, \wvec_k) \\
&= \gamma_k \lambda_k \pi(A_k|S_k) \zvec_{k-1} + I_k \nabla \estq(S_k, A_k, \wvec_k).
\end{align*}
Changing the index from $k$ to $t$, the accumulating trace update can be written as,
\begin{align*}
\zvec_t &= \gamma_t \lambda_t \pi(A_t|S_t) \zvec_{t-1} + I_t \nabla \estq(S_t, A_t, \wvec_t),
\end{align*}
leading to the incremental update for estimating $\wvec_{t+1}$.

%\iffalse
%then, the update we want at each time step is
%
%\begin{align}
%  \wvec_{t+1} = \wvec_t + \alpha I_t (G_t - \estq(S_t, A_t, \wvec_t)) \nabla \estq(S_t, A_t, \wvec_t)
%\end{align}
%
%
%Looking at only the update part, and substituting $G_t$ with the above equivalence:
%\begin{align}
%  \eta_t I_t(G_t - \estq(S_t, A_t, \wvec_t))\\
%  = \eta_t I_t \sum_{k=t}^\infty \delta_k \prod_{i=t+1}^k \gamma_i \lambda_i \pi(A_i|S_i)\\
%\end{align}
%
%First, note that for the trace 
%\begin{align}
%  \zvec_k = \gamma_k \pi(A_k|S_k) \lambda \zvec_{k-1} + I_k \nabla \estq(S_k, A_k, \wvec_k)\\
%  \zvec_k = \sum_{t=0}^k I_t \nabla \estq(S_t, A_t, \wvec_t) \prod_{i=t+1}^k \gamma_i \lambda_i \pi(A_i|S_i) 
%\end{align}
%
%Checking the equivalence of updates between the forward and backward view when the weight vector does not change
%\begin{align}
%  \sum_{t=0}^\infty \nabla \estq(S_t, A_t, \wvec) \eta_t I_t \sum_{k=t}^\infty \delta_k \prod_{i=t+1}^k \gamma_i \lambda_i \pi(A_i|S_i)\\
%  \sum_{t=0}^\infty \sum_{k=t}^\infty \nabla \estq(S_t, A_t, \wvec) \eta_t I_t  \delta_k \prod_{i=t+1}^k \gamma_i \lambda_i \pi(A_i|S_i)\\
%  \sum_{k=0}^\infty \sum_{t=0}^k \nabla \estq(S_t, A_t, \wvec)\eta_t I_t  \delta_k \prod_{i=t+1}^k \gamma_i \lambda_i \pi(A_i|S_i)\\
%  \sum_{k=0}^\infty \eta_t \delta_k  \sum_{t=0}^k I_t \nabla \estq(S_t, A_t, \wvec)  \prod_{i=t+1}^k \gamma_i \lambda_i \pi(A_i|S_i)\\
%  \sum_{k=0}^\infty \eta_t \delta_k \zvec_k
%\end{align}
%\fi

\subsection{Auto Optimizer}
We use a variant of the Autostep optimizer throughout our experiments. Adam and RMSProp are global update scaling methods and do note adapt step-sizes on a per feature basis \citep{kingma2014adam}, unlike meta descent methods like IDBD, Autostep \citep{mahmood2012tuning}, and AdaGain \citep{jacobsen2019meta}---this is critical for achieving introspective learners. Meta-descent methods like Autostep have been shown to be very effective with linear function approximation \citep{jacobsen2019meta}.
Jacobsen's AdaGain algorithm is rather complex, requiring finite differencing, whereas Auto is a simple method that works nearly as well in practice. In our own preliminary experiments, we found Adam to much less effective at tracing non-stationary learning targets, even when we adapted all three hyperparameters of the method. Finally, Auto can be seen as optimizing a meta objective for the step-size and thus is a specialization of Meta-RL to online step-size adaption in RL.

There have been attempts to apply the Autostep algorithm to TD and Sarsa [Dabney and Barto, 2012]. Auto represents another attempt to use Autostep in the reinforcement learning setting. Modifications to the Autostep algorithm are from personal communications with an author of the original work \citep{mahmood2012tuning} on how to make it more effective in practice in the reinforcement learning setting.

\begin{algorithm}[H]
	\caption{Auto Update}
	\label{alg:AutoUpdate}
  {\bfseries Input:} $\delta, \boldsymbol{\phi}, \mathbf{z}$
	\begin{algorithmic}
    \STATE $\textbf{n}_j \leftarrow \textbf{n}_j + {\tau}^{-1} \alphavec_j\abs{\boldsymbol{\phi}_j} \cdot \left(\abs{\textbf{h}_j \cdot{\delta \boldsymbol{\phi}_j}} - \textbf{n}_j \right) \quad \forall j \in \{1, ..., d\}$ 
		\FOR{\textbf{all}  $i$ such that $\boldsymbol{\phi}_i \neq 0$}
		\STATE {$\Delta \beta_i \leftarrow \text{clip}\left(-M_\Delta,\abs{\dfrac{\textbf{h}_i \delta \boldsymbol{\phi}_i}{\textbf{n}_i}}, M_\Delta \right)$} 
		\STATE {$\alphavec_i \leftarrow \text{clip}\left(\kappa, \alphavec_i e ^{\mu \Delta\beta_i}, \dfrac{1}{\abs{\boldsymbol{\phi}_i}}\right)$}
    \ENDFOR
		\IF {$\alphavec^T \textbf{z} > 1$}
		\STATE $\alphavec_i \leftarrow \min(\alphavec_i, \dfrac{1}{||\textbf{z}||_1}) \quad \forall i \ \mathbf{z}_i \neq 0$
		\ENDIF
		\STATE $\thetavec \leftarrow \thetavec + \alphavec \cdot{\delta\boldsymbol{\phi}}$
		\STATE $\textbf{h} \leftarrow \textbf{h}(1 - \alphavec \cdot{\abs{\boldsymbol{\phi}}}) + \alphavec \cdot \delta\boldsymbol{\phi}$
	\end{algorithmic}
\end{algorithm}

\pagebreak
where: \\
\vspace{-\topsep}
\begin{itemize}
\setlength{\parskip}{0pt}
\setlength{\itemsep}{0pt plus 1pt}
\item $\mu$ is the meta-step size parameter.
\item $\alphavec$ is the step sizes.
\item $\delta$ is the scalar error.
\item $\boldsymbol{\phi}$ is the feature vector.
\item $\textbf{z}$ is the  step-size truncation vector.
\item $\thetavec$ is the weight vector.
\item $\textbf{h}$ is the decaying trace.
\item $\textbf{n}$ maintains the estimate of $\abs{\textbf{h} \cdot \delta \phi}$.
\item $\tau$ is the step size normalization parameter.
\item $M_\Delta$ is the maximum update parameter of $\alphavec_i$.
\item $\kappa$ is the minimum step size.
\end{itemize}
\vspace{-\topsep}

In all experiments, $M_\Delta = 1$, $\tau = 10^4$, $\kappa = 10^{-6}$. In the reinforcement setting, $\phi$ is the eligibility trace, $\delta$ is the td error, and $\textbf{z}$ is the overshoot vector. $\textbf{z}$ is calculated as 
$\abs{\phi} \cdot \max(\abs{\phi},\abs{\textbf{x} - \gamma \textbf{x}^\prime})$, where $\textbf{x}$ is the state representation at timestep $\textit{t}$ and $\textbf{x}^\prime$ is the state representation at timestep $\textit{t}+1$.

%\clearpage
\section{Experiment Details}
\label{app:exp_details}

This section provides additional details about the experiments in the main body, and the additional experiments in this appendix. All the experiments in this work used a combined compute usage of approximately five CPU months.

\subsection{TMaze Details}\label{app:tmazedetails}

Tabular TMaze is a deterministic gridworld with four actions \{up, down, left, right\}. There are four GVFs being learned and each correspond to a goal as depicted in Figure \ref{fig_tmaze}. For GVF $i$ and the corresponding goal state $G_i$, $pi_i$, $\gamma_i$ and $\cfunc_i$ are defined as:
\begin{itemize}
    \item $\pi_i(a | s)$: deterministic policy that directs the agent towards $G_i$
    \item $\gamma_i(G_i) = 0, \  \gamma_i(s) = 0.9 \ \forall s \neq G_i \in \States$
    %\item $\cfunc_i^t(s, a, s') = C^t_i \ $ if $P(s, a, G_i) = 1$\\ $c_i^t(s, a, s') = 0$ otherwise
    \item \( \cfunc_i^t(s, a, s') =  
    			\begin{cases}
    				0 & s' \neq G_i \\
    				C^t_i & s' = G_i \\
    			\end{cases}
    		\)
\end{itemize}
where $C_i^t$ can be one of the following four different and possibly non-statioanry cumulant schedules:
\begin{itemize}
    \item Constant: $C_i^t = C_i$
    \item Distractor: $C_i^t = N(\mu_i, \sigma_i)$
    \item Drifter: $C_i^t = C_i^{t - 1} + N(\mu_i, \sigma_i), C_i^0 = 1$
\end{itemize}

As discussed in Section \ref{sec:nonstationarity-in-learning}, the cumulants of the GVFs can be stationary or non-stationary signals. The cumulant of each GVF has a non-zero value at their respective goal. In the Tabular TMaze, the top left goal is a \textit{distractor} cumulant which is an unlearnable noisy signal. The \textit{distractor} has $\mu = 1$ and $\sigma^2 = 25$. The cumulants corresponding to the lower left goal and upper right goal are \textit{constant} goals uniformly selected at the start of each run between $[-10,10]$. The cumulant corresponding to the lower left goal is a \textit{drifter} signal of $\sigma^2=0.01$ and represents a learnable non-stationary signal.

%To simulate the non-stationarity in cumulant estimates discussed in section \ref{sec:nonstationarity-in-learning}, there is a balance of stationary cumulants and non-stationary cumulants used in the experiments. In our experiments two cumulants are the \textit{static} cumulants which represent easy to learn cumulants. One of the cumulants is a \textit{distractor} cumulant providing an unlearnable noisy signal and the final cumulant is a \textit{drifter} signal which represents a learnable non-stationary signal.  

\begin{wrapfigure}{R}{0.4\textwidth}
	\includegraphics[width=0.4\textwidth]{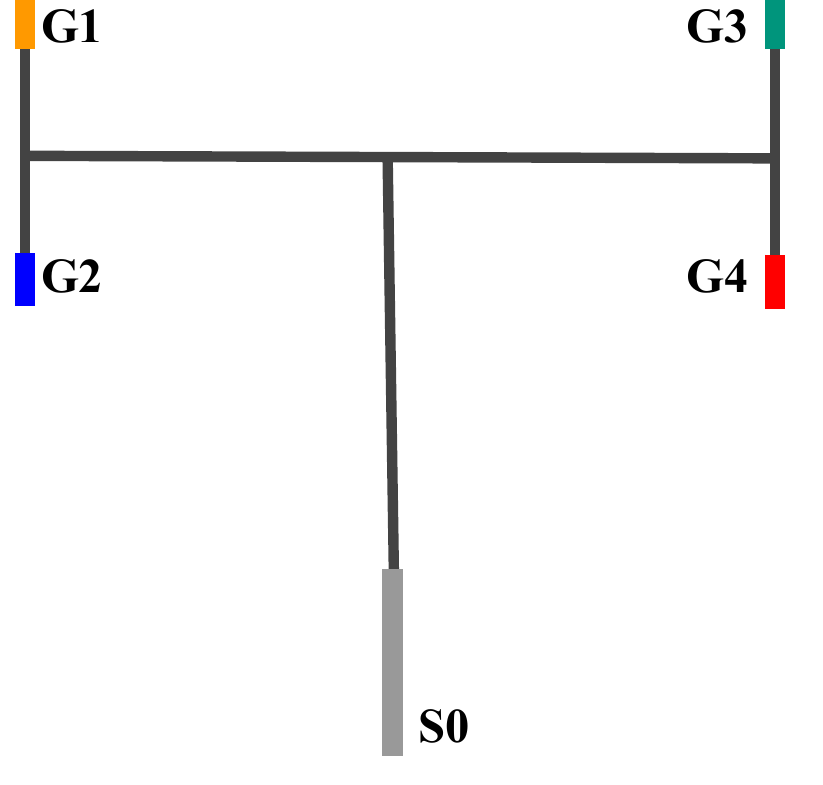}
	\caption{Continuous TMaze with the 4 GVFs. S$_0$, the grey shaded region, is the uniformly weighted starting state distribution after a goal visit.}
	\label{fig:conttmaze-env}
\end{wrapfigure}

The \emph{Continuous TMaze} follows the same design as the Tabular TMaze except it is embedded in a continuous 2D plane between 0 and 1 on both axes. Each hallway is a line with no width, allowing the agent to go along the hallway, but not perpendicular to it. 
The main vertical hallway spans between [0, 0.8] on the y-axis and is located at x of 0.5. The main horizontal hallway spans [0, 1] on the x axis and is located at y of 0.8. Finally, the two vertical side hallways span between [0.6, 1.0]. Junctions and goal locations occupy a $2\epsilon$ x $2\epsilon$ space at the end of each hallway. For example, the middle junction spans x of $0.5 \pm \epsilon$ and y of $0.8 \pm \epsilon$. The agent can take one of four actions four actions \{up, down, left, right\}. The agent moves in the corresponding direction of the action with a step size of 0.08 and noise altering movement by $\text{Uniform}(-0.01, 0.01)$. Figure \ref{fig:conttmaze-env} summarizes the environment set-up. 

%Even in this simple domain, compute costs can begin to scale and we must control for this. When learning successor features, every feature in the reward features is its own learning problem. For the Continuous TMaze, if the reward features for SF-NR or GPI were the state representation of 2 tilings of 8 tiles, then this would be essentially learning 162 GVFs per policy. For GPI, the reward features are state-action features  which results in scaling the number of reward features being learned by the number of actions. Therefore, in addition the sensitivity to poor reward features (as shown in \ref{app:reward-features}), a practical concern arises for reduced reward features from the state representation. Reduced computational cost enables more thorough sweeps and significantly more runs (all experiments in this paper have at least 30 runs per hyperparameter configuration). The sweeps on TMaze and Continuous TMaze took approximately 1 month of CPU time with the reduced reward features described in the subsection below. If the reward features were the agent state representation, then performing  the sweeps would have approximately a year of CPU time. 

\textbf{Reward features}

As discussed in the main paper, SF-NR requires a reward feature, $\phi(s,a,s')$. In the Tabular TMaze, the reward features for both the GVFs and the behavior learner are the tabular representation of $\phi(s,a,s')$. For the Continuous TMaze, the reward features using SF-NR for the GVF learners is an indicator function for if in the tuple $(s,a,s')$ is in the GVF's goal state. Since the Continuous TMaze has four goals, $\phi$ for the GVF learners is a  four dimensional vector. This is a reasonable feature vector as the reward feature vector should be related to rewarding transitions.
For GPI, it is unclear what is a rewarding transition apriori. Therefore, the reward feature is the action-feature vector of state-aggregation applied to the Continuous TMaze. This is a general  yet compact feature representation. Each line segment for the Continuous TMaze is broken up into thirds and state aggregation is applied to each part. 
%Since the reward feature vector is a state-action feature vector, $\phi_i(s,a,s') = 1$ if and only if $s$ is in the ith part and has the corresponding action {\color{red} How to explain the state-action feature vector}... do i need to explain state-action feature vectors? hmm.

\textbf{Algorithm parameters} 

For the fixed behavior experiment in Tabular TMaze, the GVFs using TB($\lambda$) learners and SF-NR learners had their meta-step size swept from $[5^{-4},...,5^{0}]$ and initial step size tested for $[0.1,1.0]$. For both TB($\lambda$) and SF-NR, the optimal meta-step size was $5^{-1}$ and initial step size of $1.0$.

For the learned behavior experiment, the behavior learner and GVF learner share the same meta-step size parameter and initial alpha. The meta-step size was swept from $[5^{-4},...,5^0]$ and initial step sizes were $[0.1,1.0]$. All four agents (GPI behavior learner with TB($\lambda$) or SF-NR GVF learners, and SARSA behavior learner with TB($\lambda$) or SF-NR GVF learners) had an optimal meta-step size of $5^{-2}$. For agents using SF-NR GVF learners, the optimal initial step size was $1.0$. For agents using TB($\lambda$) GVF learners, the initial step size was $0.1$. The behavior learner is optimistically initialized to ensure that the agent will visit each of the four goals at least once. To the best of our knowledge, no one has tried optimistic initialization with successor features before. To perform optimistic initialization for GPI, we initialized all successor features, $\psi$, estimate to be $\textbf{1}$. We initialized the immediate reward estimates, $\textbf{w}$, to the desired optimistic initialization threshold normalized by the number of reward features. We believe this to be an approximate version of optimistic initialization to allow comparisons to the SARSA agent. All behaviors used a fixed $\epsilon$ of 0.1 for the runs. The agent's performance was evaluated on the TE for the last 10\% of the runs.

For the fixed behavior experiment in Continuous TMaze, the meta-step size parameter was swept from $[5^{-3},...,5^{0}]$ and the initial step size was swept over $[0.01,0.1,0.2]$. The initial step size was then divided by the number of tilings of the tile coder to ensure proper scaling. For the fixed behavior experiment, the optimal meta-step size for SF-NR and TB($\lambda$) GVF learners was $5^{-2}$ with an initial step size of $0.2$.

For the learned behavior experiment in Continuous TMaze, the behavior learners and GVF learners shared the same meta-step size parameter and initial step size. The meta-step size parameter was swept from $[5^{-4},...,5^{0}]$ and the initial step size was swept over $[0.01,0.1,0.2]$. For GPI, the optimal initial step size for both types of GVF learners was 0.2. For SF-NR learners, the optimal meta-step size was $5^{-3}$ while being $5^{-2}$ for the TB($\lambda$) GVF learners. For the Sarsa behavior learner, the optimal meta-step size was $5^{-2}$ and the initial step size when the SF-NR GVF learners were used was $0.2$ while being $0.1$ for the TB($\lambda$) learners. The behavior learner was optimistically initialized and used an $\epsilon$ of $0.1$. The agent's performance was evaluated on the TE for the last 10\% of the runs.

Since the intrinsic reward (weight change) is $\geq 0$, the intrinsic reward is augmented with a modest $-0.01$ reward per step to encourage the agent to seek out new experiences.

\subsection{Open 2D World Details}\label{app:2dworlddetails}

\begin{wrapfigure}{R}{0.4\textwidth}
	\includegraphics[width=0.4\textwidth]{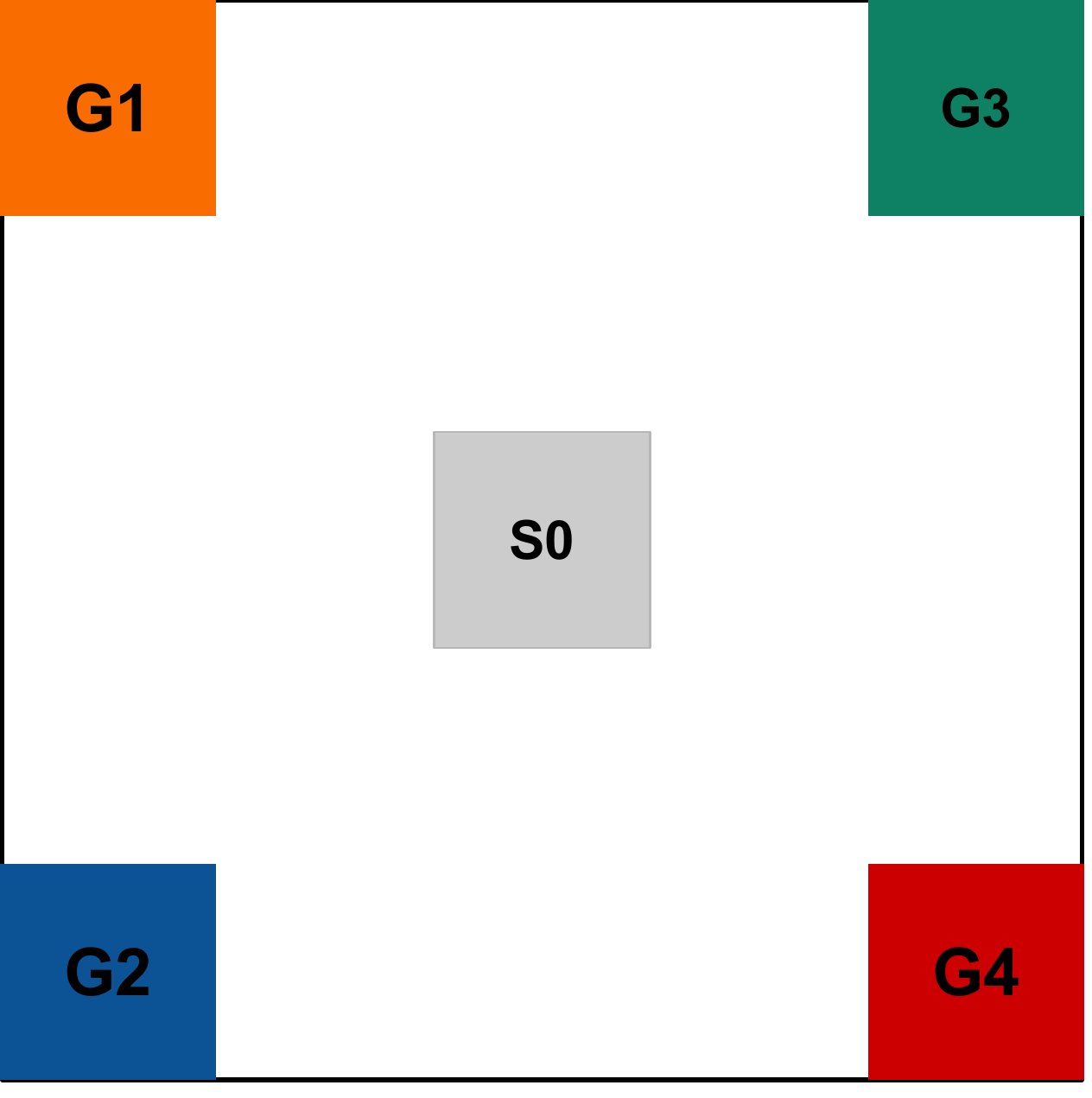}
	\caption{Open 2D World with the 4 GVFs situated at each corner. S$_0$, the grey shaded region, is the uniformly weighted starting state distribution after a goal visit.}
	\label{fig:2d-env}
\end{wrapfigure}

The Open 2D World is an open continuous grid world with boundaries defined by a square of dimensions $10\times10$, with goals in each of the four corners. The goals follow the same schedules as defined for the TMaze experiments with \textit{constants} sampled from $[-10,10]$, \textit{drifter} parameters of $\sigma^2 = 0.005$ and the initial value of $1$, and \textit{distractor} parameters of $N(\mu = 1, \sigma^2 = 1)$. On each step, the agent can select between the four compass actions. This moves the agent $0.5$ units in the chosen direction with uniform noise $[-0.1,0.1]$. A uniform $[-0.001,0.001]$ orthogonal drift is also applied. The goals are squares in the corners with a size of $1 \times 1$. The start state distributions is the center of the environment $(x,y) \in [0.45, 0.55]^2$. The summary of the environment is shown in Figure \ref{fig:2d-env}.
Similar to the TMaze variants, the GVF policies are defined as the shortest path to their respective goal. When there are multiple actions at a state that are part of a shortest path, these actions are equally weighted. The discount for the GVFs is $\gamma = 0.95$ for all states other than the goal states.

\textbf{Reward features}

The GVF reward features for SF-NR are defined similarly to the reward features in TMaze as described in Appendix \ref{app:tmazedetails}. Since there are four goals, $\phi \in \mathbf{R}^4$, where the value at $\phi_i$ is the indicator function for if $s' \in G_i$. This is a reasonable feature as it is focused on rewarding events. For the reward features of GPI, state aggregation is applied with tiles of size (2,2) and is augmented to be a state-action feature.

\textbf{Algorithm parameters}

The meta-step size for the behavior learners and the GVF learners were swept independently. The behavior learner's meta-step size was swept over $[5^{-5},...,5^0]$ while the GVF learner's meta-step size was swept over $[5^{-4},...5^0]$. An initial step size of $0.1$ scaled down by the number of tilings was used for all learners. The optimal meta-step size for the GVF learners that used ETB($\lambda$) was $5^{-3}$ and the meta-step size for the corresponding GPI learner was $5^{-1}$. Variance was an issue for learning successor features through ETB($\lambda$) so the emphasis was clipped at 1. For the agents using TB with Interest, the optimal meta-step size for the behavior learner was $5^{-1}$ and for the GVF learners was $5^{-2}$. For the method using no prior corrections, the behavior learner used a meta-step size of $5^{-4}$ and the GVF learner used a meta-step size of $5^{-2}$.

Since the intrinsic reward (weight change) is $\geq 0$, the intrinsic reward is augmented with a modest $-0.05$ reward per step to encourage the agent to seek out new experiences.

\subsection{Mountain Car Details}\label{app:mcdetails}

We use the standard mountain car environment \citep{suttonbartobook} defined through a system of equations
\begin{align*}
  A_t &\in [\text{Reverse}=-1, \text{Neutral}=0, \text{Throttle}=1]\\
  \dot{x}_{t+1} &= \dot{x}_t + 0.001 A_t - 0.0025 \cos(3x_t)\\
  x_{t+1} &= x_t + \dot{x_{t+1}}.
\end{align*}
We define two GVFs as auxiliary tasks. The first GVF receives a non-zero cumulant of value $1$ when the agent reaches the left wall. It has a discount $\gamma=0.99$ that terminates when the left wall is touched, and a policy that is learned offline to maximize the cumulant. The second GVF is similar, but receives a non-zero cumulant of $1$ when reaching the top of the hill (the typical goal state). Each policy is learned offline for 500k steps on a transformed problem of the cumulant being -1 per step with ESARSA($\lambda$) and an $\epsilon = 0.1$. This allows a denser reward signal for learning a high quality policy. Note that the final policy for maximizing this cumulant signal and the sparse reward signal are the same. The state representation used for these policies is an independent tile coder of 16 tilings with 2 tiles per dimension. The fixed policy after offline learning for each of the GVFs is greedy with respect to the learned offline action values.

\textbf{Reward features}

The reward features for the SF-NR learners are defined similarly to the reward features in the Continuous TMaze. $\phi(s,a,s')_i$ is 1 if and only if $s'$ is in the termination zone of GVF$_i$. The reward feature for GPI is a tile coder of 8 tilings with 2 tiles per dimension.

\textbf{Algorithm parameters}

The SF-NR and behavior learner are optimized with stochastic gradient descent in an online learning setting. The behavior step size and GVF step size were swept independently at $[3^{-2},...3^0]$. The values were then divided by the number of tilings to ensure proper scaling of step size. The behavior was epsilon-greedy with $\epsilon$ swept over the range $[0.1,0.3,0.5]$. For both GPI and Sarsa, $\epsilon = 0.1$ performed the best. For agents using GPI as the behavior learner, the optimal behavior step size was $3^0$ with the optimal GVF learner step size of $3^{-2}$. For the agents using a Sarsa behavior learner, the optimal parameters were a step size of $3^0$ for the behavior learner and a step size of $3^{-1}$ for the GVF learner. For the baseline agent where the behavior was random actions, the optimal GVF step size was $3^{-1}$.

Since the intrinsic reward (weight change) is $\geq 0$, the intrinsic reward is augmented with a modest $-0.01$ reward per step to encourage the agent to seek out new experiences.

%\clearpage
\section{Additional Experiments}

\subsection{Goal Visitation in Continuous TMaze}\label{app:gpi-tmaze}

%\iffalse
%\begin{wrapfigure}{R}{0.5\textwidth}
%\centering
%\includegraphics[width=0.4\textwidth]{figures/1d_tmaze/goal_visits-old.png}
%\caption{Goal visitation for GPI with PCG and Sarsa for the GVF learners using SF-NR and TB in Continuous TMaze}
%\label{fig:goal_visits_CTMaze}
%\end{wrapfigure}
%\fi

\begin{figure}
\centering
	\includegraphics[width=0.6\textwidth]{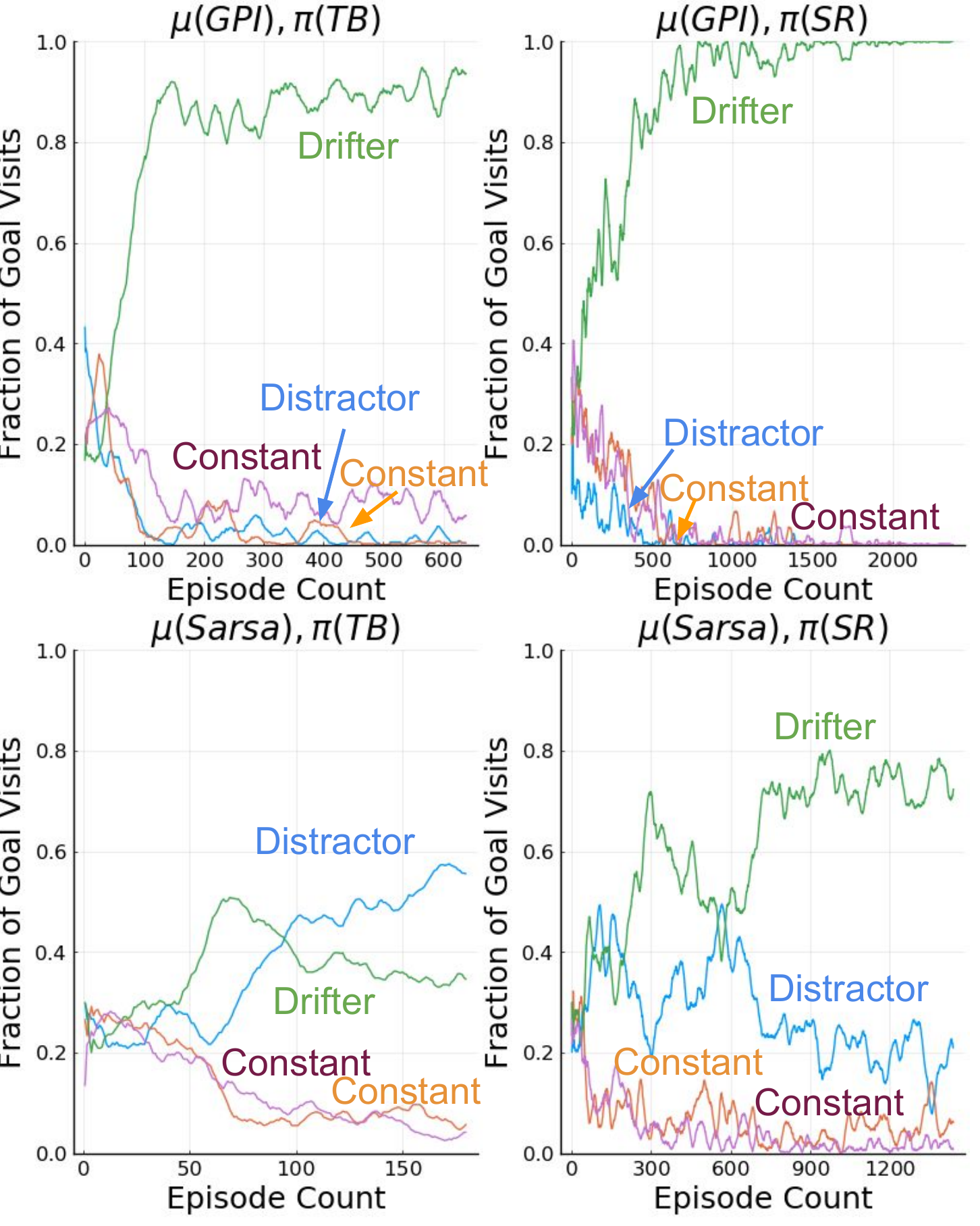}
	\caption{Goal visitation for GPI and Sarsa for the GVF learners using SF-NR and TB in Continuous TMaze. Episodes count are shown for the first N episodes up to the minimum number of episodes per run for each algorithm.}
	\label{fig:goal_visits_CTMaze}
\end{figure}

Figure \ref{fig:goal_visits_CTMaze} shows the goal visitation plots for in the Continuous TMaze for GPI and Sarsa. Using either GPI or SF-NR results in significantly faster identification of the \textit{drifter} signal and together results in the fastest identification. Prioritizing visiting the drifter is the preferred behavior as it is the only learnable signal. It is important that the agent does not get confused by the \textit{distractor} as it is not a learnable signal. 

\subsection{Goal Visitation in Tabular TMaze}\label{app:sarsa-tmaze}

%\begin{wrapfigure}{R}{0.5\textwidth}
\begin{figure}
	\centering
	\includegraphics[width=0.5\textwidth]{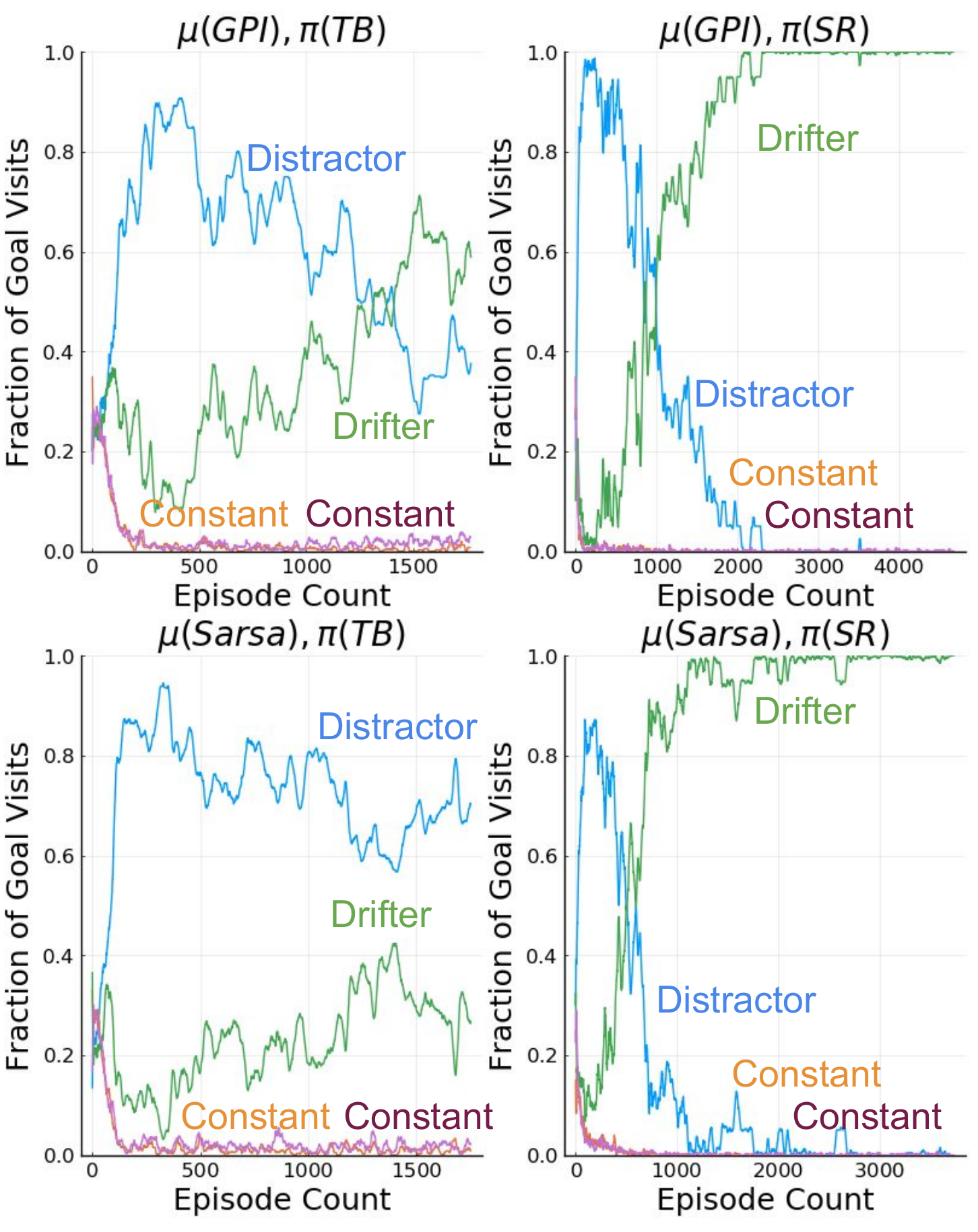}
	\caption{Goal visitation for Sarsa with the GVF learners using SF-NR and TB in Tabular TMaze.}
	\label{fig:goal_visits_TMaze}
\end{figure}
%\end{wrapfigure}

Figure \ref{fig:goal_visits_TMaze} shows the goal visitation plots for the Tabular TMaze.
% Like with GPI behavior learner in Figure~\ref{fig:visitation_tmaze}, when the behavior learner is Sarsa with SF GVF learners, increased visitation to \textit{drifter} cumulant occurs once the \textit{constant} cumulants have been learned. In contrast, with TB GVF learners, this is not so. In fact, it incorrectly focuses on the \textit{distractor} cumulant more, as compared to with GPI behavior.
%\iffalse
%\begin{figure}[h]
%	\includegraphics[width=0.8\textwidth]{figures/tabular_tmaze/goal_visitation_sarsa.png}
%	\caption{Goal visitation for Sarsa with the GVF learners using SF-NR and TB in TMaze}
%	\label{fig:goal_visits_TMaze}
%\end{figure}
%\fi

\subsection{The Effect of Generalization in the Reward Features}\label{app:reward-features}
\begin{figure}[t]
	\centering
	\includegraphics[width=0.6\textwidth]{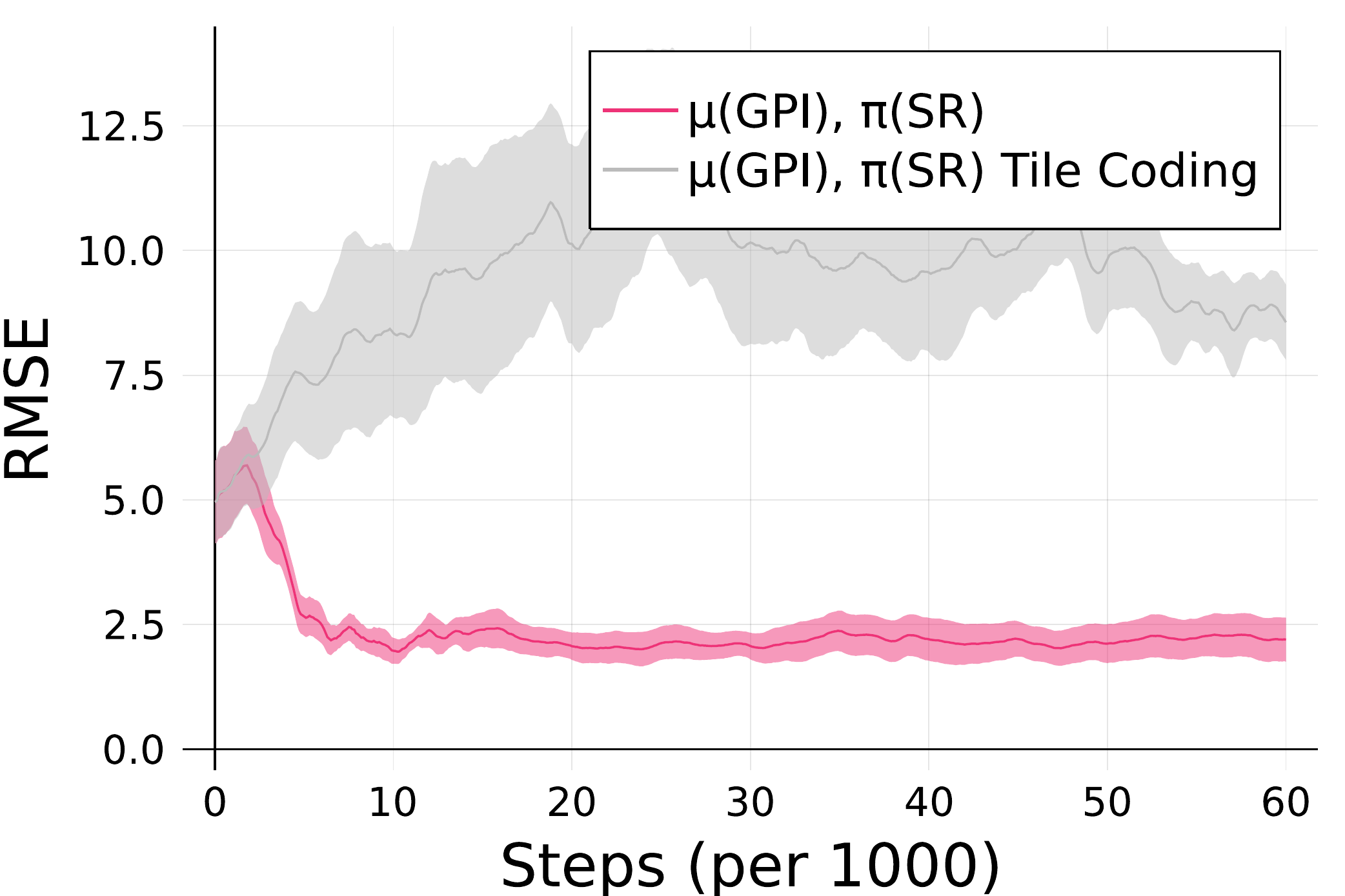}
	\caption{ GPI with the same input features, but different reward features in the Continuous TMaze environment. $\mu(GPI), \pi(SR)$ uses the reward features detailed in Appendix \ref{app:tmazedetails}. $\mu(GPI), \pi(SR) $ Tile Coding uses the reward features of a tilecoder with 8 tilings of 2 tiles.} 
	\label{fig:gpi_sensitivity}
\end{figure}

%\begin{wrapfigure}[16]{l}{0.5\textwidth}
%  \begin{centering}
%      \includegraphics[width=0.5\textwidth]{figures/GPI_sensitivity/GPI_sensitivity.pdf}
%  \end{centering}
%  \vspace{-0.4cm}
%    \caption{\label{fig:gpi_sensitivity}
%      GPI with PCG with the same input features but different reward features in the \emph{Continuous TMaze} environment. $\mu(GPI), \pi(SR)$ is with the reward features detailed in \ref{app:tmazedetails}. $\mu(GPI), \pi(SR) $ Tile Coding is with the reward features of a tilecoder with 8 tilings of 2 tiles. 
 %   }\label{gpi_sensitivity}  
%\end{wrapfigure}

GPI is sensitive to the reward feature representation that is used. Figure~\ref{fig:gpi_sensitivity} shows what happens when the reward features of 8 tilings with 2x2 tiles are used in the Continuous TMaze for GPI. The agents are swept over the same interval of hyperparameters as described in Appendix ~\ref{app:tmazedetails} for the Continuous TMaze. When this reward feature was used, the optimal meta-step size was $5^{-3}$. The initial step size was $0.2$ which was then scaled down by the number of tilings. GPI, with this reward features, was unable to learn an effective policy to reduce the TE in the Continuous TMaze. The tile coded reward feature representation has approximately three times the number of features than the handcrafted feature representation, yet it results in much worse performance. This highlights the need for the reward features to be learnable by an algorithm rather than being predefined.

%Figure \ref{fig:cumulantexample} show an example of the different cumulant schedules over time. Each cumulant schedule is desflect a different set of difficulties when generating intrinsic rewards, with the static cumulant schedule as the baseline. The drifter cumulant schedule is a non-stationary cumulant that requires the agent to often visit the goal state to learn the true value. The distractor cumulant schedule asks the agent to identify when an accurate prediction is already obtained despite noisigned to ree in the signal.

%\begin{figure}[h]
%    \includegraphics[width=6cm]{figures/tmaze.pdf}
%    \centering
%    \caption{\textbf{Tabular TMaze} with 4 prediction problems. The arrow shows the policy toward the goal state $G_i$ for each of the 4 prediction problems.}
%    \label{fig:tmaze}
%\end{figure}

%
%\begin{figure}[h]
%    \includegraphics[width=8cm]{figures/cumulant_example.pdf}
%    \centering
%    \caption{\textbf{Example of the three different cumulant schedules}: the \textit{static}, \textit{distractor}, and \textit{drifter}cumulant schedules.}
%    \label{fig:cumulantexample}
%\end{figure}

%%% Local Variables:
%%% mode: latex
%%% TeX-master: "main"
%%% End:

\end{document}